\documentclass{article}
\usepackage{iclr2027_conference,times}

\usepackage{amsmath,amsfonts,bm}

\def\eqref#1{equation~\ref{#1}}

\def\1{\bm{1}}

\DeclareMathAlphabet{\mathsfit}{\encodingdefault}{\sfdefault}{m}{sl}
\SetMathAlphabet{\mathsfit}{bold}{\encodingdefault}{\sfdefault}{bx}{n}

\usepackage[pagebackref=false,breaklinks=true,colorlinks,bookmarks=false,citecolor=citecolor,linkcolor=linkcolor]{hyperref}  
\usepackage[capitalize,noabbrev]{cleveref}
\usepackage{url}
\usepackage{graphicx}
\usepackage{booktabs}
\usepackage{amsmath,amssymb}
\usepackage{colortbl}
\usepackage{multirow}
\usepackage{enumitem}
\usepackage{xspace}
\usepackage{array}
\newcolumntype{C}{r}
\newcolumntype{F}{r}
\newcolumntype{L}{r}
\newcolumntype{E}{r}
\newcolumntype{A}{r}

\crefname{figure}{Fig.}{Figs.}
\Crefname{figure}{Figure}{Figures}

\crefname{table}{Table}{Tables}
\Crefname{table}{Table}{Tables}

\crefname{section}{Sec.}{Secs.}
\Crefname{section}{Section}{Sections}

\crefname{equation}{Eq.}{Eqs.}
\Crefname{equation}{Equation}{Equations}

\crefname{appendix}{Appendix}{Appendices}
\Crefname{appendix}{Appendix}{Appendices}

\definecolor{citecolor}{HTML}{0071BC}
\definecolor{linkcolor}{HTML}{ED1C24}

\renewcommand{\cite}{\citep}

\newcommand{\eg}{\textit{e.g.}\xspace}

\graphicspath{{figures/}}

\title{Continual Concept Erasure in Diffusion Models by Suppressing Cross-Edit Interference}

\author{Yongliang Wu$^{1}$, Haori Lu$^{1}$, Jinqi Luo$^{2}$, Wei Cao$^{1}$, Xingyu Zhu$^{3}$\thanks{Corresponding author}, Yaoyao Liu$^{1}$ \\$^{1}$ University of Illinois Urbana-Champaign \\ $^{2}$ University of Pennsylvania \\ $^{3}$ National University of Singapore \\
}

\iclrfinalcopy 
\begin{document}

\maketitle

\begin{abstract}
Concept erasure removes copyright-protected, privacy-sensitive, or otherwise undesirable concepts from pretrained text-to-image diffusion models to support content governance and compliance. As erasure requests arrive over time, models must remove new targets without undoing prior erasures. Existing methods do not constrain interference across edits: residual perturbations outside the retain set interact and accumulate, degrading unrelated generations and sometimes collapsing previously erased targets into noise. We propose \textbf{CEASE} (\textbf{C}ontinual \textbf{E}rasure via \textbf{A}daptive \textbf{S}ubspace \textbf{E}diting), a training-free method that imposes two subspace constraints on a closed-form solver. CEASE adds the token representation of the shared replacement to the solver's invariance matrix and, when interference is detected, projects the current update onto the orthogonal complement of dominant output directions extracted from cumulative past updates. A closed-form decomposition attributes the accumulated interference to repeated activation of the shared replacement and overlap between successive update directions, showing that the two constraints suppress these respective sources. Across continual erasure of celebrities, artistic styles, and instances, CEASE achieves the most consistent erase--preserve trade-off, while existing methods either degrade general generation or insufficiently erase targets.\footnote{Project page: \url{https://continual-erasure.cvmlgroup.web.illinois.edu/}}

\end{abstract}

\section{Introduction}
\label{sec:intro}

Concept erasure aims to remove copyright-protected, privacy-sensitive, or otherwise undesirable concepts from pretrained text-to-image (T2I) diffusion models, making it an important tool for content governance and compliance~\cite{gandikota2023erasing, lu2024mace, liu2026image, liu2025rethinking}. In practice, new erasure targets arrive over time~\cite{lee2026continual,george2025distill}, extending concept erasure from one-shot to continual editing: the model must remove new targets without allowing later updates to undo prior erasures or degrade generation of unrelated concepts.

Prior studies report that retained model utility declines when erasure requests are processed sequentially~\cite{lee2026continual,george2025distill}. In our experiments, sequentially applying standard erasure methods~\cite{gandikota2024unified,lu2024mace,gong2024reliable,li2026speed} reveals two related failure modes. First, general generation quality declines on prompts unrelated to any erased concept. Second, outputs for some erased targets become incoherent instead of reflecting the prescribed replacement (\eg ``a person'' for a celebrity). Each edit explicitly protects only a finite retain set~\cite{gandikota2024unified,gong2024reliable,li2026speed} and can therefore perturb embeddings outside it~\cite{wuerkaixi2025adaptive,thakral2025continual}.

\begin{figure}[t]
\centering
\includegraphics[width=\linewidth]{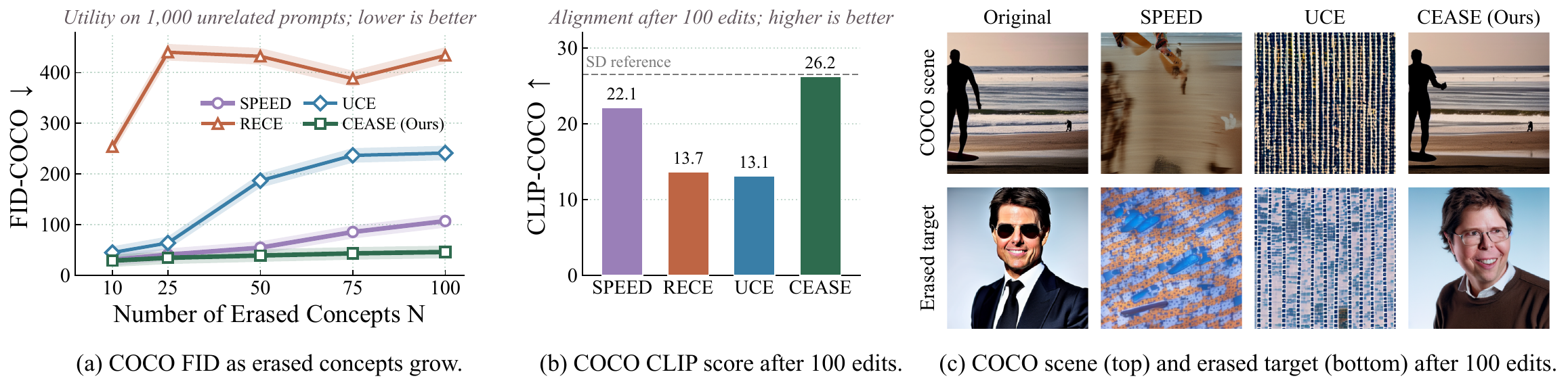}
\caption{\textbf{Utility loss under continual celebrity erasure.} \textbf{(a)} COCO FID on 1,000 unrelated prompts as the number of erased concepts grows from 10 to 100; lower is better. \textbf{(b)} COCO CLIP score on the same prompts after 100 edits; the dashed line marks the unedited model. \textbf{(c)} Generations after 100 edits for an unrelated COCO scene (top) and the erased target ``Tom Cruise'' (bottom).}
\label{fig:main}
\end{figure}

These perturbations interact and accumulate across successive updates, making it difficult for existing methods~\cite{li2026speed,gong2024reliable,gandikota2024unified} to confine the effects of erasure to the target concepts. As shown in Fig.~\ref{fig:main}(a), COCO FID on unrelated prompts rises as the number of edits increases~\cite{lin2014microsoft}. This exposes a limitation of existing preservation constraints in the continual setting: protecting a finite retain set at each step does not reliably prevent subsequent updates from cumulatively degrading general generation. Beyond image quality, this degradation also weakens the model's ability to follow prompt semantics: \Cref{fig:main}(b) shows that, after 100 edits, the CLIP score on the same prompts falls well below that of the unedited model. This reduced ability to follow prompt semantics may also affect the outcomes of previous erasures, because the replacement outputs for earlier targets depend on the same continually updated weights. \Cref{fig:main}(c) illustrates both effects in the baseline models: they corrupt an unrelated scene and produce noise for the erased celebrity, failing to generate the prescribed generic portrait.

To suppress the accumulation of interference in continual concept erasure, we propose \textbf{CEASE} (\textbf{C}ontinual \textbf{E}rasure via \textbf{A}daptive \textbf{S}ubspace \textbf{E}diting), a training-free method based on two subspace constraints. CEASE adds the token representation of the shared replacement to the invariance matrix of the closed-form solver, preventing subsequent updates from modifying this representation. It also extracts the dominant output directions from cumulative past updates and, when interference is detected, projects the current update onto their orthogonal complement. Theoretically, a closed-form decomposition attributes the accumulated interference to repeated activation of the shared replacement and overlap between successive update directions, and shows that the two designs suppress the respective sources. Empirically, CEASE maintains stable COCO FID throughout the edit sequence and a COCO CLIP score close to the unedited reference after 100 edits, as shown in Fig.~\ref{fig:main}(a) and (b). Across continual erasure tasks involving celebrities, artistic styles, and instances, CEASE achieves the most consistent erase--preserve trade-off. On style erasure, CEASE reduces COCO FID by 73\% relative to the strongest prior method. Our contributions are summarized as follows:

\begin{itemize}[leftmargin=*]
    \item We formalize how interference accumulates across edits in continual concept erasure, and derive a closed-form decomposition that attributes it to repeated activation of the shared replacement and overlap between successive update directions.
    \item We introduce two training-free subspace constraints: one keeps the token representation of the shared replacement invariant, while the other steers the current update away from the dominant output directions of past edits, thereby suppressing the two interference sources.
    \item We evaluate continual erasure of celebrities, artistic styles, and instances across multiple diffusion backbones and assess both erasure and preservation; CEASE achieves the best erase--preserve balance among the compared methods while keeping retained concepts recognizable at rates close to the reference model.
\end{itemize}

\section{Related Work}
\label{sec:related}

\noindent \textbf{Concept Erasure in Diffusion Models.} Text-to-image diffusion models~\cite{zhang2023adding,ma2023generating,ji2026ac3s} trained on uncurated web data can reproduce copyrighted styles, recognizable identities, and other undesired content, and concept erasure removes such targets from a pretrained model after training. Existing methods fall into three families. Training-based methods fine-tune the model with a concept-suppression objective and a preservation regularizer~\cite{gandikota2023erasing,zhang2024forget,kumari2023ablating,huang2024receler,lyu2024one,li2024safegen,kim2024race,zhao2024advanchor,wu2025unlearning}, with a separate strand hardening the erasure against adversarial prompts~\cite{fan2024salun,heng2023selective,pham2024robust,park2024direct}; their common cost is that each new target requires another fine-tuning loop. Editing-based methods instead update cross-attention projections through a closed-form least-squares solution, from TIME~\cite{orgad2023editing} and UCE~\cite{gandikota2024unified} to MACE~\cite{lu2024mace}, RECE~\cite{gong2024reliable}, and SPEED~\cite{li2026speed}; they edit the model in seconds, but their preservation depends on a finite retain set. Non-parametric methods leave the weights untouched and intervene at sampling time~\cite{schramowski2023safe,das2024espresso,rando2022red}. All three families are typically evaluated with the full target set known before editing; our study instead evaluates repeated application to an already-edited model.

\noindent \textbf{Sequential Unlearning.} Sequential unlearning is closely related to continual learning and online learning~\cite{liu2020mnemonics,liu2021adaptive,liu2021rmm,liu2023online,liu2023continual,liu2024wakening,li2019online,li2020online,li2021online,luo2023class,fischer2024inemo,zhang2023continual}. While continual learning and online learning typically focus on acquiring new knowledge without forgetting, sequential unlearning poses the complementary challenge of repeatedly removing targeted knowledge without causing unintended forgetting of retained concepts. In deployment, takedown demands and right-to-be-forgotten obligations arrive as a stream. For language models, the locate-then-edit paradigm~\cite{meng2022locating,meng2022mass} degrades under repeated application, and mitigations include drift-restraining regularizers~\cite{ma2025perturbation,gu2024model}, external edit memories~\cite{hartvigsen2023aging,wang2024wise}, and projections onto the null space of preserved keys~\cite{fang2025alphaedit}; these approaches operate on factual associations, where edits do not share a common redirection target. For diffusion models, Selective Amnesia~\cite{heng2023selective} first cast concept forgetting as a continual-learning problem, and later work adds concept-aware masking with distillation~\cite{li2025sculpting}, gradient projection against semantically related concepts~\cite{lee2026continual}, and multi-objective distillation across deletion requests~\cite{george2025distill}; these methods suppress the degradation as a monolithic quantity without analyzing where it comes from. In contrast, we approach continual erasure from the perspective of cross-edit interference: we trace the failure to the interference that successive edits accumulate on unprotected prompts, and design our method to suppress this interference.

\section{Analysis and Method}
\label{sec:method}

Section~\ref{sec:method:prelim} defines the continual erasure problem and the base closed-form update. Section~\ref{sec:method:analysis} then explains why update noise accumulates under sequential editing. Guided by the two sources identified in this analysis, Sec.~\ref{sec:method:steer} presents CEASE.

\subsection{Problem Setup}
\label{sec:method:prelim}

Concept erasure for T2I diffusion models edits the cross-attention value projection $\mathbf{W} \in \mathbb{R}^{d_1 \times d_0}$, which maps a frozen CLIP text embedding $\boldsymbol{c} \in \mathbb{R}^{d_0}$ to the value vector $\mathbf{W} \boldsymbol{c} \in \mathbb{R}^{d_1}$, via an additive update $\mathbf{W}' = \mathbf{W} + \boldsymbol{\Delta}$. We specify an erasure set $\mathbf{E} = \{ \boldsymbol{c}_1^{(i)} \}_{i=1}^{N_E}$ to remove and a retain set $\mathbf{R} = \{ \boldsymbol{c}_0^{(i)} \}_{i=1}^{N_R}$ to preserve, pair each target with an anchor $\boldsymbol{c}_*^{(i)}$ that keeps generic semantics but discards target-specific information, and stack these column-wise as $\mathbf{C}_1, \mathbf{C}_* \in \mathbb{R}^{d_0 \times N_E}$ and $\mathbf{C}_0 \in \mathbb{R}^{d_0 \times N_R}$.

\noindent \textbf{Base Closed-Form Update.} Closed-form methods~\cite{gandikota2024unified, li2026speed} choose $\boldsymbol{\Delta}$ by solving a weighted least-squares problem that balances two requirements:
\begin{equation}
    \boldsymbol{\Delta}_{\text{erase}} \;=\; \arg\min_{\boldsymbol{\Delta}} \; \underbrace{\bigl\| (\mathbf{W} + \boldsymbol{\Delta}) \mathbf{C}_1 - \mathbf{W} \mathbf{C}_* \bigr\|^2}_{e_1: \text{ redirect targets}} \;+\; \underbrace{\bigl\| \boldsymbol{\Delta} \mathbf{C}_0 \bigr\|^2}_{e_0: \text{ preserve non-targets}}.
\label{eq:uce}
\end{equation}
$e_1$ redirects each target's output toward its anchor value $\mathbf{W} \boldsymbol{c}_*^{(i)}$, while $e_0$ preserves non-target outputs. Because both terms depend on the same update, adding targets can increase the preservation residual~\cite{li2026speed}. Following SPEED, the base solver reduces this residual by right-projecting the update through a near-null projector $\mathbf{P}$ of $\mathbf{C}_0^\top$. The applied update $\boldsymbol{\Delta}^{\star}\mathbf{P}$ therefore satisfies $(\boldsymbol{\Delta}^{\star}\mathbf{P})\mathbf{C}_0\approx\mathbf{0}$, but this guarantee covers only the span of the finite retain set. Appendix~\ref{app:nullspace} provides the projector construction.

\noindent \textbf{Continual Concept Erasure.} The continual setting replaces the joint solve with $N$ updates applied one after another. Starting from the pretrained weights, each step solves for a new update against the running weights and applies it:
\begin{equation}
    \mathbf{W}^{(0)} \;=\; \mathbf{W}, \qquad \mathbf{W}^{(i)} \;=\; \mathbf{W}^{(i-1)} + \boldsymbol{\Delta}_i, \qquad \boldsymbol{\Delta}_{\mathrm{sum}} \;=\; \sum_{i=1}^N \boldsymbol{\Delta}_i,
\label{eq:sequential}
\end{equation}
where step $i$ handles a single target and its anchor, with the retain set held fixed throughout.

\subsection{Why Update Noise Accumulates}
\label{sec:method:analysis}

The base solver protects only embeddings in the span of $\mathbf{C}_0$. Each edit can therefore introduce an unintended change to the value-projection output of an uncovered prompt, and these changes accumulate over the sequence.

\noindent \textbf{Measuring Accumulated Update Noise.} We quantify the accumulated change directly at the value-projection output. Consider an arbitrary test embedding $\boldsymbol{c}$ outside the retain set. After all $N$ edits, the cross-attention value becomes $\bigl( \mathbf{W} + \sum_{i \leq N} \boldsymbol{\Delta}_i \bigr) \boldsymbol{c}$, whose magnitude is bounded by the triangle inequality:
\begin{equation}
    \Bigl\| \bigl( \mathbf{W} + \sum_{i \leq N} \boldsymbol{\Delta}_i \bigr) \boldsymbol{c} \Bigr\|_2 \;\leq\; \bigl\| \mathbf{W} \boldsymbol{c} \bigr\|_2 \;+\; \Bigl\| \sum_{i \leq N} \boldsymbol{\Delta}_i \boldsymbol{c} \Bigr\|_2.
\label{eq:drift_bound}
\end{equation}
The first term $\| \mathbf{W} \boldsymbol{c} \|_2$ is fixed, so the change induced by editing is governed by $\| \sum_i \boldsymbol{\Delta}_i \boldsymbol{c} \|_2$. Let $e$ denote the single edit intended to affect $\boldsymbol{c}$; for a prompt unrelated to every erased target, we set $\boldsymbol{\Delta}_e\boldsymbol{c}=\mathbf{0}$. The desired reference is therefore $\|\boldsymbol{\Delta}_e\boldsymbol{c}\|_2^2$. We define the accumulated update noise as the difference between the squared magnitude of the full update and this reference:
\begin{equation}
    \mathrm{noise}_e(\boldsymbol{c}) \;=\; \Bigl\| \sum_{i \leq N} \boldsymbol{\Delta}_i \boldsymbol{c} \Bigr\|_2^2 \;-\; \bigl\| \boldsymbol{\Delta}_e \boldsymbol{c} \bigr\|_2^2,
\label{eq:noise_def}
\end{equation}
a signed residual over the sequence. Positive values indicate that the full sequence produces a larger update than the intended edit alone, whereas negative values indicate cancellation among updates.

To identify why this noise grows, we use the rank-at-most-one structure of the single-target, single-token update. At step $i$, the closed-form solver acts on one target $\boldsymbol{c}_1^{(i)}$ and anchor $\boldsymbol{c}_*^{(i)}$; left- and right-multiplication by its projection matrices preserves the rank bound of the target outer product. We may therefore write $\boldsymbol{\Delta}_i = \boldsymbol{\alpha}_i \boldsymbol{\beta}_i^\top$, where $\boldsymbol{\alpha}_i$ is the output-space influence direction and $\boldsymbol{\beta}_i$ determines how strongly an arbitrary $\boldsymbol{c}$ activates the edit, since $\boldsymbol{\Delta}_i \boldsymbol{c} = (\boldsymbol{c}^\top \boldsymbol{\beta}_i)\, \boldsymbol{\alpha}_i$. Substituting this into Eq.~\ref{eq:noise_def} gives:
\begin{equation}
    \mathrm{noise}_e(\boldsymbol{c}) \;=\; \sum_{\substack{i,j \\ (i,j)\ne(e,e)}} \bigl( \boldsymbol{c}^\top \boldsymbol{\beta}_i \bigr) \bigl( \boldsymbol{\alpha}_i^\top \boldsymbol{\alpha}_j \bigr) \bigl( \boldsymbol{\beta}_j^\top \boldsymbol{c} \bigr).
\label{eq:noise_exp}
\end{equation}
\noindent \textbf{Decomposing Accumulated Update Noise.} We split any uncovered prompt into its component along the anchor and its orthogonal remainder:
\begin{equation}
\boldsymbol{c} \;=\; s\,\boldsymbol{c}_* + \boldsymbol{c}_\perp, \qquad s = \frac{\langle \boldsymbol{c}_*, \boldsymbol{c} \rangle}{\|\boldsymbol{c}_*\|_2^2}, \quad \boldsymbol{c}_\perp \perp \boldsymbol{c}_*.
\label{eq:prompt_split}
\end{equation}
Each activation then splits into an anchor part and an off-anchor part:
\begin{equation}
\boldsymbol{\beta}_i^\top \boldsymbol{c} \;=\; s\,(\boldsymbol{\beta}_i^\top \boldsymbol{c}_*) \;+\; \boldsymbol{\beta}_i^\top \boldsymbol{c}_\perp .
\label{eq:act_split}
\end{equation}
For compactness, let $a_i=\boldsymbol{\beta}_i^\top\boldsymbol{c}_*$, $r_i=\boldsymbol{\beta}_i^\top\boldsymbol{c}_\perp$, and $g_{ij}=\langle\boldsymbol{\alpha}_i,\boldsymbol{\alpha}_j\rangle$, and let $\sum_{i,j}'$ exclude $(i,j)=(e,e)$. Substituting Eq.~\ref{eq:act_split} into Eq.~\ref{eq:noise_exp} and combining the two symmetric cross terms gives:
\begin{equation}
\begin{gathered}
\mathrm{noise}_e(\boldsymbol{c})
    = \mathcal{N}_{\mathrm{anc}}(\boldsymbol{c})
    + \mathcal{N}_{\mathrm{off}}(\boldsymbol{c})
    + \mathcal{N}_{\mathrm{cross}}(\boldsymbol{c}), \\[2pt]
\mathcal{N}_{\mathrm{anc}}(\boldsymbol{c})=s^2\sum_{i,j}'a_i a_j g_{ij}, \qquad
\mathcal{N}_{\mathrm{off}}(\boldsymbol{c})=\sum_{i,j}'r_i r_j g_{ij}, \qquad
\mathcal{N}_{\mathrm{cross}}(\boldsymbol{c})=2s\sum_{i,j}'a_i r_j g_{ij}.
\end{gathered}
\label{eq:noise_split}
\end{equation}
The anchor channel $\mathcal{N}_{\mathrm{anc}}$ is driven by the anchor activations $\boldsymbol{\beta}_i^\top \boldsymbol{c}_*$: every edit uses the same fixed direction $\boldsymbol{c}_*$, so these contributions stack across steps for prompts with a nonzero anchor component $s$. The off-anchor channel $\mathcal{N}_{\mathrm{off}}$ is driven by the off-anchor activations and the influence overlap $\boldsymbol{\alpha}_i^\top \boldsymbol{\alpha}_j$. The cross channel $\mathcal{N}_{\mathrm{cross}}$ couples the two via the anchor activations. $\mathcal{N}_{\mathrm{anc}}$ and $\mathcal{N}_{\mathrm{cross}}$ both vanish once $\boldsymbol{\beta}_i^\top \boldsymbol{c}_* = 0$, while the terms coupling different edits in $\mathcal{N}_{\mathrm{off}}$ vanish once $\boldsymbol{\alpha}_i^\top \boldsymbol{\alpha}_j = 0$ for $i \ne j$. These conditions motivate the AIC and HOC constraints of CEASE.

\subsection{CEASE: Suppressing Update Noise}
\label{sec:method:steer}

\begin{figure}[t]
\centering
\includegraphics[width=\linewidth]{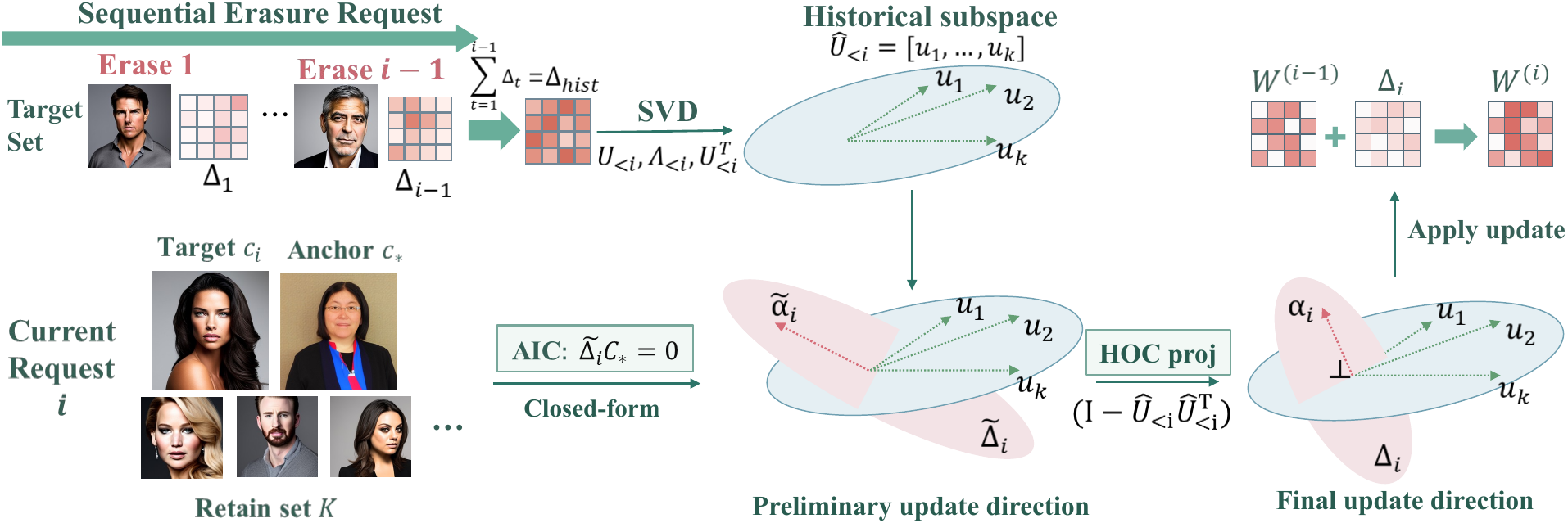}
\caption{Overview of CEASE for continual concept erasure. AIC preserves the shared anchor during the closed-form solve. HOC extracts dominant output directions from cumulative historical edits via SVD and, when needed, projects the new update onto their orthogonal complement before applying it to the current model.}
\label{fig:method}
\end{figure}

The decomposition in Eq.~\ref{eq:noise_split} identifies two ideal conditions under which only the $O(N)$ diagonal remains: $\boldsymbol{\beta}_i^\top \boldsymbol{c}_* = 0$ for all $i$, and $\boldsymbol{\alpha}_i^\top \boldsymbol{\alpha}_j = 0$ for $i \ne j$. CEASE targets these conditions with two components, Anchor Invariance and Historical Orthogonality. As illustrated in Fig.~\ref{fig:method}, AIC protects the shared anchor while HOC projects each new update away from dominant historical edit directions. Both reuse the existing closed-form update and require no retraining.

\noindent{\textbf{Anchor Invariance Constraint (AIC).}} The SPEED solver already contains an invariance matrix $\mathbf{K}_2$ formed from the empty-prompt representation, but its default construction does not include the task anchor $\boldsymbol{c}_*$. AIC appends the anchor token embedding to this matrix:
\begin{equation}
\mathbf{K}_2^{\mathrm{AIC}} \;=\; [\mathbf{K}_2,\,\boldsymbol{c}_*], \qquad
\boldsymbol{\Delta}_i\mathbf{K}_2^{\mathrm{AIC}} \;=\; \mathbf{0}.
\label{eq:aic}
\end{equation}
With the zero invariance regularizer used in our experiments, the closed-form correction enforces $\boldsymbol{\Delta}_i\boldsymbol{c}_*=\mathbf{0}$ up to numerical precision. For the rank-at-most-one factorization, this implies $\boldsymbol{\beta}_i^\top\boldsymbol{c}_*=0$ whenever $\boldsymbol{\alpha}_i\ne\mathbf{0}$. Under this ideal constraint, the anchor and cross groups vanish:
\begin{equation}
\mathcal{N}_{\mathrm{anc}}(\boldsymbol{c}) \;=\; 0, \qquad \mathcal{N}_{\mathrm{cross}}(\boldsymbol{c}) \;=\; 0, \qquad \forall \boldsymbol{c},
\label{eq:aic_zero}
\end{equation}
leaving the off-anchor interaction for the next component to address.

\noindent{\textbf{Historical Orthogonality Constraint (HOC).}} Under the ideal AIC constraint in Eq.~\ref{eq:aic_zero}, only the off-anchor channel $\mathcal{N}_{\mathrm{off}}$ remains. Separating it by step index:
\begin{equation}
\mathcal{N}_{\mathrm{off}}(\boldsymbol{c}) = \underbrace{\sum_{i\ne e} (\boldsymbol{\beta}_i^\top \boldsymbol{c}_\perp)^2\,\|\boldsymbol{\alpha}_i\|^2}_{\text{diagonal},~O(N)} \;+\; \underbrace{\sum_{i\ne j} (\boldsymbol{\beta}_i^\top \boldsymbol{c}_\perp)\,(\boldsymbol{\alpha}_i^\top \boldsymbol{\alpha}_j)\,(\boldsymbol{\beta}_j^\top \boldsymbol{c}_\perp)}_{\text{cross},~O(N^2)},
\label{eq:off_split}
\end{equation}
The diagonal terms capture single-step effects, while the cross terms scale with the influence overlap $\boldsymbol{\alpha}_i^\top \boldsymbol{\alpha}_j$ between edits. HOC reduces this overlap when the accumulated historical update substantially affects the current target, thereby mitigating interference between successive edits.

A naive way to enforce this orthogonality would keep all past influence vectors $\{\boldsymbol{\alpha}_j\}_{j<i}$ in memory and project each new $\boldsymbol{\alpha}_i$ onto their orthogonal complement, which grows linearly in $N$. We avoid this overhead by exploiting the rank-1 structure of every closed-form update. Since $\boldsymbol{\Delta}_j = \boldsymbol{\alpha}_j \boldsymbol{\beta}_j^\top$, the cumulative historical edit is:
\begin{equation}
\boldsymbol{\Delta}_{\mathrm{hist}} \;=\; \sum_{j<i} \boldsymbol{\Delta}_j \;=\; \sum_{j<i} \boldsymbol{\alpha}_j \boldsymbol{\beta}_j^\top
\label{eq:delta_hist}
\end{equation}
Its column space is contained in $\mathrm{span}\{\boldsymbol{\alpha}_j\}_{j<i}$. It therefore provides a compact, observable proxy for historical output directions, although cancellations can make it a strict subset. We obtain its dominant left singular directions from:
\begin{equation}
    \{\mathbf{U}_{<i}, \boldsymbol{\Lambda}_{<i}, \mathbf{U}_{<i}^\top\} \;=\; \mathrm{SVD}\bigl( \boldsymbol{\Delta}_{\mathrm{hist}} \boldsymbol{\Delta}_{\mathrm{hist}}^\top \bigr).
\label{eq:hoc_svd}
\end{equation}
Using the dominant left singular vectors $\widehat{\mathbf{U}}_{<i}$ from Eq.~\ref{eq:hoc_svd}, HOC projects the AIC-constrained update $\tilde{\boldsymbol{\Delta}}_i$ onto the orthogonal complement of the historical update subspace. Accordingly, $\boldsymbol{\Pi}_i=\mathbf{I}-\widehat{\mathbf{U}}_{<i}\widehat{\mathbf{U}}_{<i}^{\top}$ when HOC is applied, and $\boldsymbol{\Pi}_i=\mathbf{I}$ otherwise. CEASE updates the model as:
\begin{equation}
\boldsymbol{\Delta}_i=\boldsymbol{\Pi}_i\tilde{\boldsymbol{\Delta}}_i,
\qquad
\mathbf{W}^{(i)}=\mathbf{W}^{(i-1)}+\boldsymbol{\Delta}_i.
\label{eq:cease_update}
\end{equation}

\section{Experiments}
\label{sec:experiments}

\subsection{Experimental Setup}
\label{sec:exp:setup}

\noindent \textbf{Baselines and Benchmarks.} We evaluate on three concept erasure benchmarks adapted from MACE~\cite{lu2024mace} and SPEED~\cite{li2026speed}, covering celebrities, artistic styles, and object or character instances. Each benchmark pairs 100 erased concepts with 100 disjoint retained concepts and uses five prompt templates per concept. We compare two groups of baselines: standard erasure methods, comprising the closed-form SPEED~\cite{li2026speed}, UCE~\cite{gandikota2024unified}, and RECE~\cite{gong2024reliable} and the training-based ESD~\cite{gandikota2023erasing} and ConAbl~\cite{kumari2023ablating}, and continual-unlearning methods, comprising CUIG-GradProj~\cite{lee2026continual} and Sculpting Memory~\cite{li2025sculpting}.

\noindent \textbf{Metrics.} We report CLIP score (CS)~\citep{hessel2021clipscore} for image--prompt alignment, and FID~\citep{heusel2017gans} and LPIPS~\citep{zhang2018unreasonable} for the distribution-level and image-level distance from the unedited backbone's generations, on each of the erase set, the retain set, and 1{,}000 MS-COCO captions. On the erase and retain sets we add one task-matched fourth metric. For celebrities, Acc is the recognition accuracy of an open-ended VLM judge (Qwen2-VL-7B)~\cite{wang2024qwen2} asked to name the person in the image; for instances, Acc is the CLIP zero-shot top-1 accuracy against the 200 concept candidates; both should drop on the erase set and stay high on the retain set. For styles, the LAION aesthetic score (Aes)~\citep{schuhmann2022laion} is a no-reference quality rating on a 1--10 scale that should stay near the unedited reference.

\noindent \textbf{Implementation Details.} The main experiments use Stable Diffusion v1.4 with DPM-Solver sampling and classifier-free guidance 7.5~\cite{rombach2021highresolution}. The per-task anchors are ``person'' for celebrities, ``art'' for styles, and ``object'' for instances. All experiments run on a single H100 GPU. Additional implementation details, including hyperparameters, are provided in Appendix~\ref{app:impl}.

\subsection{Main Results}
\label{sec:exp:main}

\begin{table}[t]
\centering
\caption{Celebrity erasure after continually erasing 100 celebrities on SD v1.4. Bold and \underline{underline} mark the best and second-best edited results.}
\vspace{0.1cm}
\label{tab:celebrity}
\resizebox{\hsize}{!}{
\renewcommand{\arraystretch}{1.20}
\begin{tabular}{l|CFLA|CFLA|CFL}
\toprule
 & \multicolumn{4}{c|}{Erase set} & \multicolumn{4}{c|}{Retain set} & \multicolumn{3}{c}{MS-COCO} \\
 & {CS $\downarrow$} & {FID $\uparrow$} & {LPIPS $\uparrow$} & {Acc $\downarrow$} & {CS $\uparrow$} & {FID $\downarrow$} & {LPIPS $\downarrow$} & {Acc $\uparrow$} & {CS $\uparrow$} & {FID $\downarrow$} & {LPIPS $\downarrow$} \\
\midrule
SD v1.4 & 30.95 & 0.0 & 0.000 & 88.0 & 31.66 & 0.0 & 0.000 & 70.6 & 26.52 & 0.0 & 0.000 \\
\midrule
ESD & 21.64 & 61.6 & 0.545 & 23.0 & 21.31 & 59.3 & 0.544 & 12.9 & 21.96 & 61.5 & 0.540 \\
UCE & \bfseries 13.55 & 272.8 & 0.768 & \bfseries 0.0 & 14.10 & 242.7 & 0.768 & 0.0 & 13.15 & 241.4 & 0.773 \\
ConAbl & 15.79 & 74.1 & 0.540 & \bfseries 0.0 & 15.85 & 73.0 & 0.537 & 0.0 & 24.26 & 50.7 & 0.476 \\
RECE & 16.51 & \bfseries 568.3 & \bfseries 0.974 & \bfseries 0.0 & 16.27 & 573.1 & 0.972 & 0.0 & 16.28 & 616.5 & 1.002 \\
SPEED & 18.38 & 137.2 & 0.635 & {\underline{0.3}} & {\underline{30.28}} & {\underline{20.8}} & {\underline{0.383}} & {\underline{54.4}} & 22.14 & 107.2 & 0.566 \\
CUIG-GradProj & {\underline{15.28}} & {\underline{333.8}} & {\underline{0.846}} & \bfseries 0.0 & 15.03 & 336.0 & 0.844 & 0.0 & 14.15 & 303.9 & 0.849 \\
Sculpting Memory & 16.84 & 101.5 & 0.606 & 1.0 & 17.31 & 93.2 & 0.597 & 0.3 & {\underline{25.26}} & {\underline{47.5}} & {\underline{0.429}} \\
\midrule
Ours & 21.54 & 39.9 & 0.478 & 3.5 & \bfseries 31.59 & \bfseries 12.3 & \bfseries 0.271 & \bfseries 65.9 & \bfseries 26.24 & \bfseries 46.2 & \bfseries 0.388 \\
\bottomrule
\end{tabular}
}
\end{table}

We compare CEASE with all baselines after the full 100-step continual erasure sequence, and report the quantitative results on the celebrity, style, and instance benchmarks in Tables~\ref{tab:celebrity}, \ref{tab:style}, and \ref{tab:instance}, which we discuss in turn below. Across all three tasks, CEASE achieves the most favorable balance between erasure and preservation: it substantially alters the erased concepts while keeping the retained concepts and unrelated generation close to the unedited model, whereas the baselines that erase more deeply do so by corrupting the retain set or MS-COCO prompts as well. Results on the Stable Diffusion v1.5 and v2.1 backbones appear in Appendices~\ref{app:sd15} and \ref{app:sd21}, with the same conclusions.

\noindent\textbf{Erasure of Celebrities.} As shown in Table~\ref{tab:celebrity}, continually applying SPEED severely degrades general generation, with quality on both the retain set and COCO deteriorating as edits accumulate. CEASE performs best among the edited methods on nearly all retain and COCO preservation metrics, and the VLM judge almost never names the erased celebrities while still recognizing most of the retained ones. UCE and RECE obtain the lowest erase-set CLIP scores, but their retain-set scores collapse to a similar degree, indicating prompt-independent corruption rather than selective removal; ESD, in contrast, leaves a noticeable fraction of the erased identities recognizable.

\begin{table}[t]
\centering
\vspace{-0.2cm}
\caption{Style erasure after continually erasing 100 artistic styles on SD v1.4. Bold and \underline{underline} mark the best and second-best edited results.}
\label{tab:style}
\vspace{0.1cm}
\resizebox{\hsize}{!}{
\renewcommand{\arraystretch}{1.20}
\begin{tabular}{l|CFLE|CFLE|CFL}
\toprule
 & \multicolumn{4}{c|}{Erase set} & \multicolumn{4}{c|}{Retain set} & \multicolumn{3}{c}{MS-COCO} \\
 & {CS $\downarrow$} & {FID $\uparrow$} & {LPIPS $\uparrow$} & {Aes $\uparrow$} & {CS $\uparrow$} & {FID $\downarrow$} & {LPIPS $\downarrow$} & {Aes $\uparrow$} & {CS $\uparrow$} & {FID $\downarrow$} & {LPIPS $\downarrow$} \\
\midrule
SD v1.4 & 28.31 & 0.0 & 0.000 & 6.01 & 28.20 & 0.0 & 0.000 & 5.99 & 26.52 & 0.0 & 0.000 \\
\midrule
ESD & 19.98 & 85.2 & 0.617 & 5.36 & 20.12 & {\underline{84.2}} & 0.618 & 5.39 & 23.50 & 56.7 & 0.517 \\
UCE & 17.28 & 250.0 & 0.772 & 4.27 & 15.83 & 210.5 & 0.768 & 3.93 & 13.97 & 228.0 & 0.751 \\
ConAbl & 20.27 & 127.1 & 0.577 & {\underline{5.52}} & {\underline{20.16}} & 125.3 & {\underline{0.581}} & {\underline{5.51}} & 22.98 & 57.7 & {\underline{0.516}} \\
RECE & \bfseries 12.75 & \bfseries 388.7 & \bfseries 0.916 & 3.13 & 12.41 & 365.3 & 0.909 & 3.20 & 13.79 & 371.6 & 0.904 \\
SPEED & {\underline{14.94}} & 249.1 & 0.736 & 5.07 & 15.75 & 197.4 & 0.695 & 5.20 & 13.73 & 188.8 & 0.682 \\
CUIG-GradProj & 15.11 & {\underline{316.6}} & {\underline{0.828}} & 3.52 & 14.74 & 317.1 & 0.827 & 3.52 & 23.56 & 85.9 & 0.642 \\
Sculpting Memory & 21.43 & 117.8 & 0.601 & 5.49 & {\underline{21.43}} & 109.2 & 0.596 & 5.47 & {\underline{25.14}} & \bfseries 42.5 & \bfseries 0.403 \\
\midrule
Ours & 23.43 & 64.8 & 0.518 & \bfseries 5.91 & \bfseries 25.66 & \bfseries 38.7 & \bfseries 0.446 & \bfseries 5.96 & \bfseries 25.43 & {\underline{50.3}} & {\underline{0.473}} \\
\bottomrule
\end{tabular}
}
\end{table}

\begin{table}[t]
\centering
\caption{Instance erasure after continually erasing 100 instances on SD v1.4. Bold and \underline{underline} mark the best and second-best edited results.}
\label{tab:instance}
\vspace{0.1cm}
\resizebox{\hsize}{!}{
\renewcommand{\arraystretch}{1.20}
\begin{tabular}{l|CFLA|CFLA|CFL}
\toprule
 & \multicolumn{4}{c|}{Erase set} & \multicolumn{4}{c|}{Retain set} & \multicolumn{3}{c}{MS-COCO} \\
 & {CS $\downarrow$} & {FID $\uparrow$} & {LPIPS $\uparrow$} & {Acc $\downarrow$} & {CS $\uparrow$} & {FID $\downarrow$} & {LPIPS $\downarrow$} & {Acc $\uparrow$} & {CS $\uparrow$} & {FID $\downarrow$} & {LPIPS $\downarrow$} \\
\midrule
SD v1.4 & 28.02 & 0.0 & 0.000 & 80.2 & 28.46 & 0.0 & 0.000 & 78.6 & 26.52 & 0.0 & 0.000 \\
\midrule
ESD & 20.08 & 65.1 & 0.621 & 28.9 & 20.22 & 60.1 & 0.604 & 27.7 & 21.82 & {\underline{64.2}} & 0.555 \\
UCE & 15.65 & {\underline{319.8}} & 0.790 & 1.2 & 16.13 & 271.7 & 0.771 & 0.2 & 14.20 & 276.3 & 0.750 \\
ConAbl & \bfseries 15.20 & 179.7 & 0.653 & 1.8 & 15.13 & 176.7 & 0.651 & 2.5 & 19.13 & 89.9 & 0.575 \\
RECE & 16.22 & \bfseries 438.6 & \bfseries 0.891 & \bfseries 1.0 & 15.90 & 433.5 & 0.897 & 0.0 & 13.46 & 427.6 & 0.881 \\
SPEED & 18.21 & 85.0 & 0.644 & 5.7 & {\underline{26.28}} & {\underline{25.1}} & {\underline{0.479}} & {\underline{66.2}} & {\underline{22.80}} & 78.0 & {\underline{0.531}} \\
CUIG-GradProj & 16.13 & 249.4 & {\underline{0.883}} & {\underline{1.1}} & 15.65 & 250.6 & 0.882 & 0.5 & 14.20 & 229.3 & 0.892 \\
Sculpting Memory & {\underline{15.24}} & 156.2 & 0.691 & 2.7 & 15.30 & 157.6 & 0.689 & 2.3 & 22.00 & 70.0 & 0.538 \\
\midrule
Ours & 21.08 & 55.4 & 0.605 & 11.8 & \bfseries 27.15 & \bfseries 20.8 & \bfseries 0.451 & \bfseries 71.0 & \bfseries 25.26 & \bfseries 62.1 & \bfseries 0.490 \\
\bottomrule
\end{tabular}
}
\end{table}

\noindent\textbf{Erasure of Styles.} Table~\ref{tab:style} shows the style task to be SPEED's most severe failure, whereas CEASE remains best among the edited methods on all retain metrics and on COCO CLIP score. Sculpting Memory attains lower COCO FID, but at the cost of a substantially more degraded retain set. Importantly, the preservation gain of CEASE does not come from leaving the erased styles unchanged: it still lowers the erase-set CLIP score substantially, and its erase-set aesthetic score stays at the reference level, showing that the redirected outputs remain coherent paintings, whereas RECE and CUIG-GradProj produce aesthetically collapsed images.

\noindent\textbf{Erasure of Instances.} As reported in Table~\ref{tab:instance}, the instance task exhibits SPEED's mildest yet still substantial degradation, and CEASE again leads all retain and COCO preservation metrics while clearly altering the erased instances. ConAbl and RECE erase more deeply by erase-set CLIP score, but their collapsed retain scores expose the corresponding loss of selectivity. The accuracy columns make this explicit: CEASE keeps retain accuracy closest to the unedited reference while reducing erase accuracy to a low level, whereas the deepest-erasing baselines lose both together.

\begin{figure}[t]
\centering
\includegraphics[width=\linewidth]{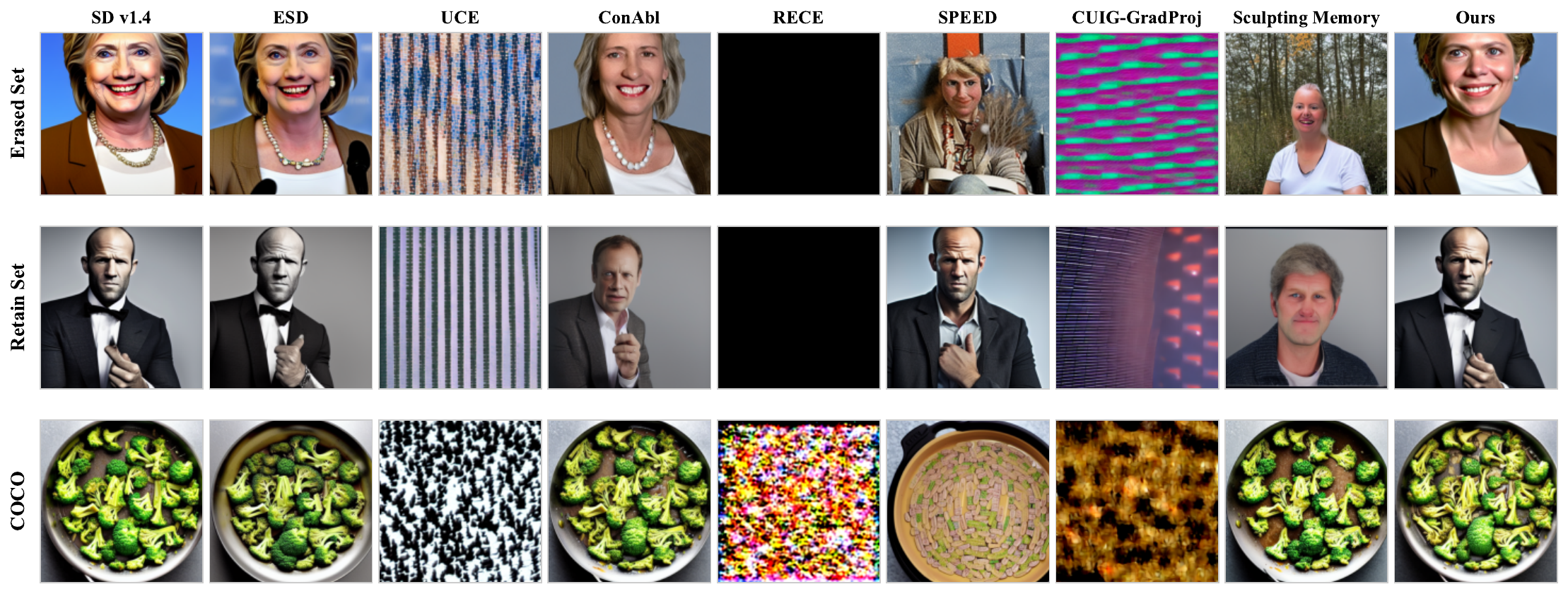}
\caption{Qualitative comparison after the 100-step continual erasure, on the erased target Hillary Clinton, the retained concept Jason Statham, and an unrelated MS-COCO prompt. CEASE redirects the target to a coherent anchor substitute while preserving the retained concept and the COCO scene.}
\label{fig:qual_main}
\end{figure}

\noindent\textbf{Qualitative Results.} \Cref{fig:qual_main} compares all methods on the erased target ``Hillary Clinton'', the retained concept ``Jason Statham'', and an unrelated MS-COCO prompt. RECE, UCE, and CUIG-GradProj collapse all three categories alike to black or noise; SPEED and ConAbl distort the erased identity and, like Sculpting Memory, replace the retained one; ESD leaves the erased target fully recognizable. In contrast, CEASE redirects the target to a coherent anchor substitute while leaving the retained concept and the COCO scene visually intact. More visualization results on the celebrity, style, and instance tasks, as well as on the Stable Diffusion v1.5 and v2.1 backbones, are provided in Appendix~\ref{app:qual_v14}, we observe the same pattern in those settings.

\subsection{Ablation Study}
\label{sec:exp:ablation}

\begin{table}[t]
\centering
\caption{Ablation of AIC and HOC on the celebrity task. Combining both modules improves retained-concept preservation and general generation quality. Bold marks the best results.}
\label{tab:ablation}
\vspace{0.1cm}
\resizebox{\hsize}{!}{
\renewcommand{\arraystretch}{1.20}
\begin{tabular}{cc|CFLA|CFLA|CFL}
\toprule
\multicolumn{2}{c|}{Module} & \multicolumn{4}{c|}{Erase set} & \multicolumn{4}{c|}{Retain set} & \multicolumn{3}{c}{MS-COCO} \\
AIC & HOC & {CS $\downarrow$} & {FID $\uparrow$} & {LPIPS $\uparrow$} & {Acc $\downarrow$} & {CS $\uparrow$} & {FID $\downarrow$} & {LPIPS $\downarrow$} & {Acc $\uparrow$} & {CS $\uparrow$} & {FID $\downarrow$} & {LPIPS $\downarrow$} \\
\midrule
$\times$ & $\times$ & \bfseries 18.38 & \bfseries 137.2 & \bfseries 0.635 & \bfseries 0.3 & 30.28 & 20.8 & 0.383 & 54.4 & 22.14 & 107.2 & 0.566 \\
$\times$ & $\checkmark$ & 20.84 & 57.8 & 0.528 & 1.5 & 31.44 & 13.5 & 0.298 & 63.7 & 25.54 & 59.7 & 0.457 \\
$\checkmark$ & $\times$ & 21.41 & 43.1 & 0.485 & 3.5 & 31.54 & 12.9 & 0.284 & 65.2 & 26.23 & 48.6 & 0.399 \\
\midrule
$\checkmark$ & $\checkmark$ & 21.54 & 39.9 & 0.478 & 3.5 & \bfseries 31.59 & \bfseries 12.3 & \bfseries 0.271 & \bfseries 65.9 & \bfseries 26.24 & \bfseries 46.2 & \bfseries 0.388 \\
\bottomrule
\end{tabular}
}
\end{table}

\begin{figure}[t]
\centering
\includegraphics[width=\linewidth]{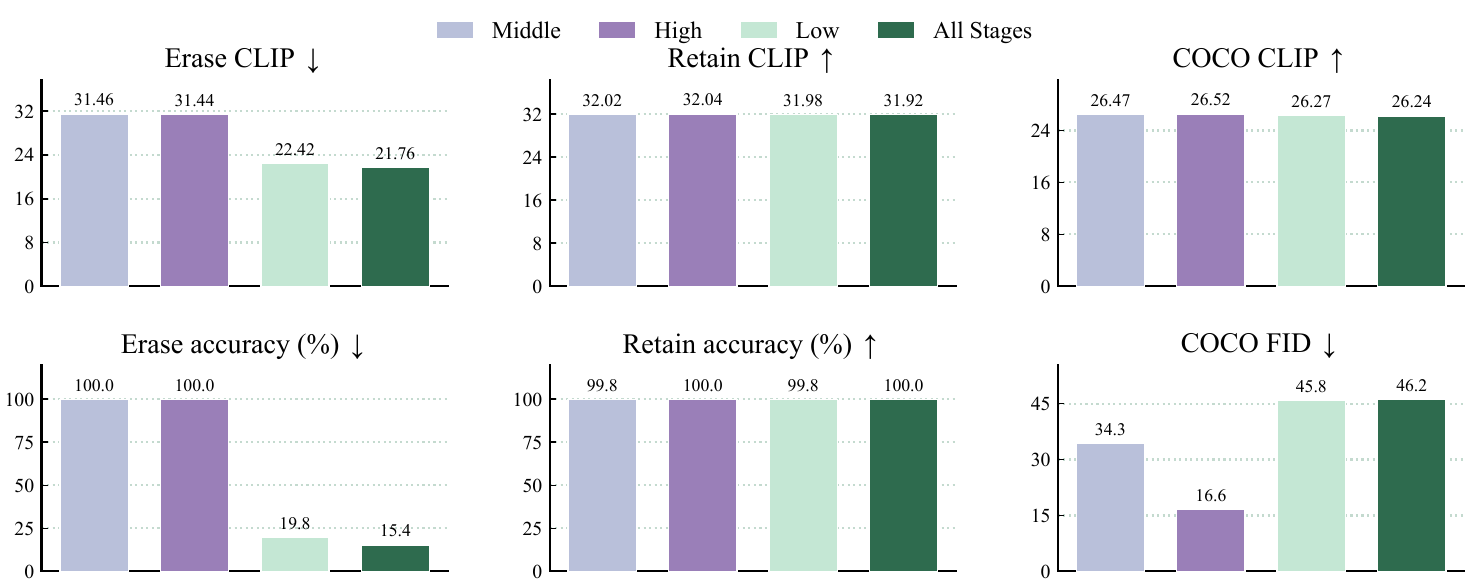}
\caption{Cross-attention layer-scope ablation on the 100-step celebrity sequence. We compare four editing scopes using erase and retain CLIP scores and recognition accuracy, together with COCO CLIP score and FID. Arrows indicate the preferred direction.}
\label{fig:layer_ablation}
\end{figure}

\begin{figure}[t]
\centering
\includegraphics[width=\linewidth]{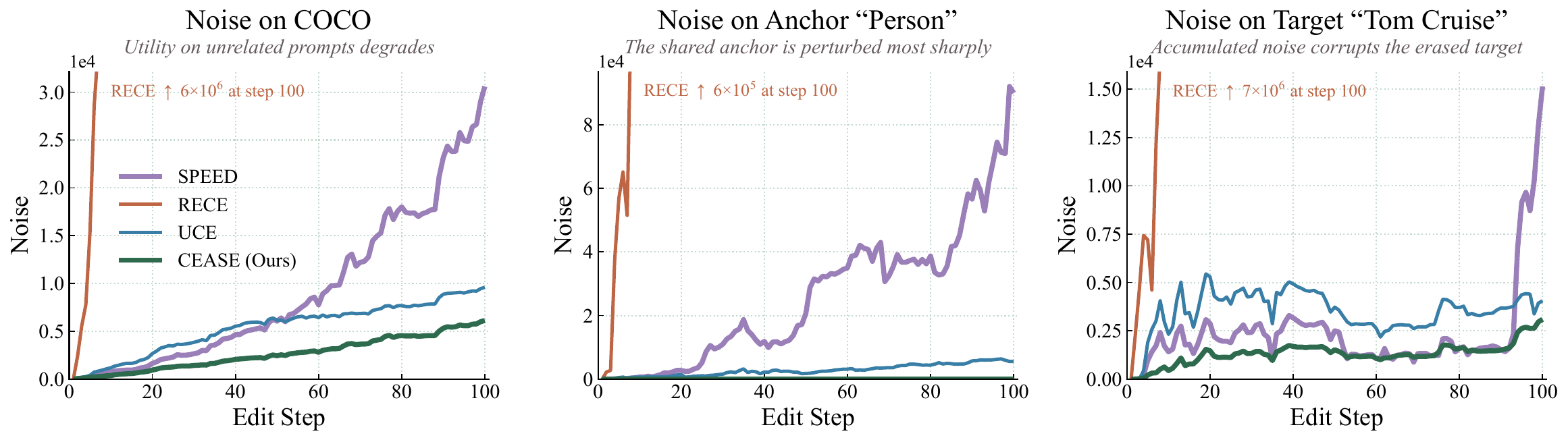}
\caption{Cross-edit interference over 100 celebrity erasures, comparing SPEED, RECE, UCE, and CEASE. The three panels track noise on unrelated COCO prompts, the shared replacement ``person'', and the erased target ``Tom Cruise'' as successive edits accumulate.}
\label{fig:ablation_scaling}
\end{figure}

\noindent \textbf{Effect of AIC and HOC.} Table~\ref{tab:ablation} isolates the two components on the celebrity task. Each component alone already recovers much of the preservation lost by the baseline, with AIC contributing the larger gain, and their combination performs best on all retain and COCO metrics. The erase-set columns show the accompanying trade-off: the full method changes the erased celebrities less strongly than the baseline, redirecting them to coherent anchor substitutes rather than removing them more weakly, so the preservation gain does not come at the cost of failed erasure.

\noindent \textbf{Cross-Attention Layer Scope.} \Cref{fig:layer_ablation} compares four layer scopes on the celebrity sequence. The middle and high-resolution scopes preserve non-target behavior but fail to erase, whereas the low-resolution and all-stage variants erase effectively with similar preservation. All Stages achieves the strongest erasure overall while retaining comparable utility, so we use it as the default.

\noindent \textbf{Cross-Edit Interference over the Sequence.} \Cref{fig:ablation_scaling} traces the value-projection interference of SPEED, RECE with one adversarial epoch per request, UCE, and CEASE over the 100-step celebrity sequence. For SPEED, interference grows steadily on unrelated COCO prompts, on the shared replacement, and on the erased target, and the target curve rises sharply in the final edits. RECE exceeds the displayed range within the first ten edits and ends several orders of magnitude higher, consistent with its COCO FID in \cref{fig:main}(a). CEASE stays below SPEED on all three diagnostics; on the shared replacement it remains near zero, which is the direct effect of AIC. UCE is the exception that shows the limits of this diagnostic: its final interference on the shared replacement is three orders of magnitude above CEASE, while its unscaled COCO and target values end slightly lower despite far worse generation quality (the displayed UCE curve is scaled by a factor of two and ends above CEASE), so value-projection interference alone does not rank preservation across methods. \Cref{fig:scaling_supplement} reports the COCO CLIP score and the target-averaged diagnostic at five checkpoints; relative to SPEED, CEASE keeps a higher CLIP score and lower FID at every checkpoint. The underlying measurements use saved checkpoints from complete sequential runs; UCE is displayed with the stated post-processing scale; the projection diagnostic is interpreted separately from generation quality (Appendix~\ref{app:noise_scope}).

In addition, we analyze the robustness of CEASE to the erasure order in Appendix~\ref{app:order}, and compare its computational efficiency with all baselines in Appendix~\ref{app:runtime}.

\section{Conclusion}
\label{sec:conclusion}

We studied continual concept erasure, where deletion requests arrive one after another and each edit builds on all previous ones. We traced the severe degradation of existing methods in this regime to accumulating cross-edit interference, and proposed CEASE, a training-free framework that protects the shared anchor and steers each new update away from the directions perturbed by past edits. Experiments on erasing celebrities, artistic styles, and instances across multiple backbones show that CEASE achieves the most consistent erase--preserve trade-off.

\noindent \textbf{Limitations.} First, our experiments focus on the Stable Diffusion family, and extending CEASE to more recent architectures such as FLUX is a natural next step. Second, this work instantiates the proposed constraints in a closed-form solver, and generalizing them to training-based erasure methods is left to future work. Third, our sequences contain up to 100 edits; evaluating substantially longer streams is conceptually straightforward but computationally expensive, since every additional step requires regenerating and re-evaluating thousands of images.

\subsection*{Acknowledgments}
This research is supported by the National Artificial Intelligence Research Resource Pilot Awards NAIRR250199, NAIRR260019, and NAIRR260077; AMD University Program’s AI \& HPC Cluster; NVIDIA Academic Grant Program; Lambda's Research Grant; and Google Cloud Research Credits Program. Computational resources are also provided by Delta and DeltaAI at the National Center for Supercomputing Applications (NCSA) through ACCESS allocations CIS250012, CIS250816, and CIS251188.

\bibliography{iclr2027_conference}
\bibliographystyle{iclr2027_conference}

\newpage
\appendix
\section{Appendix}
\subsection{Implementation Details}
\label{app:impl}

ESD~\cite{gandikota2023erasing} uses the ESD-x variant, fine-tuning the cross-attention projections under the negative-guidance objective for 200 Adam iterations per step at learning rate $1\times10^{-4}$ with guidance strength 1.0, and training latents are generated at guidance 3.0 over 50 DDIM steps.

ConAbl~\cite{kumari2023ablating} fine-tunes the cross-attention weights toward its own anchor-prompt prediction, with prior-loss weight 1.0 and 100 Adam iterations per concept at learning rate $1\times10^{-4}$.

UCE~\cite{gandikota2024unified} applies one closed-form least-squares update per step to the key and value projections, with erase scale 1.0, preserve scale 1.0, and regularization $\lambda=0.5$.

RECE~\cite{gong2024reliable} alternates a closed-form value edit with the derivation of an adversarial target embedding, in the replace regime with 3 adversarial epochs, erase scale 1.0, preserve scale 0.1, closed-form $\lambda=0.1$, and adversarial regularization 0.1.

SPEED~\cite{li2026speed} redirects each target to the anchor while projecting the update onto the null space of an augmented retain set; its invariance matrix uses the empty-prompt representation, and CEASE's AIC additionally inserts the task anchor. Both use null-space threshold 0.1, retain scale 0.05, $\lambda=0$, 10 augmented retain samples per concept, and the official multi-concept configuration without the importance-based retain filter.

CUIG-GradProj~\cite{lee2026continual} adds the official semantic-aware gradient-projection regularizer to ConAbl with 200 auxiliary prompts, fine-tuning the key and value projections with batch size 4, learning rate $2\times10^{-6}$ before the official scaling, 500 warmup steps, and weight decay $10^{-2}$, for 2,000 iterations per request (1,000 for style).

Sculpting Memory~\cite{li2025sculpting} combines a dynamic parameter mask with distillation from the preceding model, using 100 generated images per path, batch size 2, 10 epochs, learning rate $3\times10^{-6}$, and a cosine OneCycle schedule with 50 warmup steps; the mask updates every 100 steps with fraction 0.5, and the unlearning and distillation scales are 0.25.

\subsection{Near-Null Projector}
\label{app:nullspace}

For completeness, SPEED constructs the retain projector from the eigendecomposition of the retain Gram matrix,
\begin{equation}
    \mathbf{C}_0\mathbf{C}_0^\top = \mathbf{U}\boldsymbol{\Lambda}\mathbf{U}^\top,
    \qquad
    \mathbf{P}=\widehat{\mathbf{U}}\widehat{\mathbf{U}}^\top,
\end{equation}
where $\widehat{\mathbf{U}}$ contains the eigenvectors whose eigenvalues are below a threshold $\tau$. An update right-projected as $\boldsymbol{\Delta}^{\star}\mathbf{P}$ therefore lies in the approximate null space of the retain embeddings and satisfies $(\boldsymbol{\Delta}^{\star}\mathbf{P})\mathbf{C}_0\approx\mathbf{0}$. We use $\tau=0.1$ in all experiments, following SPEED.

\subsection{Backbone Generalization to Stable Diffusion v1.5}
\label{app:sd15}

We repeat the full 100-step sequence on Stable Diffusion v1.5, a second CLIP-L backbone, for all tasks and baselines, and report the results in Tables~\ref{tab:sd15_celebrity}, \ref{tab:sd15_style}, and \ref{tab:sd15_instance}. The erase--preserve pattern of the main results transfers: CEASE improves the retain and COCO preservation metrics over SPEED on all three tasks while still altering the erased targets, and its aesthetic scores on styles stay at the reference level. The baselines that erase more deeply pay for it on preservation: ConAbl attains a lower COCO FID on celebrity and style but largely destroys the retained concepts, and UCE and RECE erase deepest with collapsed retain sets.

\begin{table}[h]
\centering
\caption{Celebrity erasure after continually erasing 100 celebrities on SD v1.5. Bold and \underline{underline} mark the best and second-best edited results.}
\label{tab:sd15_celebrity}
\resizebox{\hsize}{!}{
\renewcommand{\arraystretch}{1.20}
\begin{tabular}{l|CFLA|CFLA|CFL}
\toprule
 & \multicolumn{4}{c|}{Erase set} & \multicolumn{4}{c|}{Retain set} & \multicolumn{3}{c}{MS-COCO} \\
 & {CS $\downarrow$} & {FID $\uparrow$} & {LPIPS $\uparrow$} & {Acc $\downarrow$} & {CS $\uparrow$} & {FID $\downarrow$} & {LPIPS $\downarrow$} & {Acc $\uparrow$} & {CS $\uparrow$} & {FID $\downarrow$} & {LPIPS $\downarrow$} \\
\midrule
SD v1.5 & 31.13 & 0.0 & 0.000 & 88.8 & 31.92 & 0.0 & 0.000 & 70.6 & 26.43 & 0.0 & 0.000 \\
\midrule
ESD & 28.74 & 57.4 & 0.413 & 79.2 & 28.86 & 56.5 & 0.415 & {\underline{58.2}} & 24.82 & 46.1 & 0.446 \\
UCE & \bfseries 13.08 & 363.3 & 0.818 & \bfseries 0.0 & 13.63 & 293.9 & 0.795 & 0.0 & 13.10 & 294.4 & 0.788 \\
ConAbl & 20.04 & 98.5 & 0.454 & 2.0 & 21.35 & 92.2 & 0.447 & 1.8 & {\underline{25.72}} & \bfseries 37.3 & \bfseries 0.291 \\
RECE & 16.51 & \bfseries 564.4 & \bfseries 0.977 & \bfseries 0.0 & 16.27 & 572.0 & 0.975 & 0.0 & 16.28 & 610.4 & 1.002 \\
SPEED & 18.55 & 186.3 & 0.586 & {\underline{0.6}} & {\underline{30.54}} & {\underline{50.8}} & {\underline{0.348}} & 53.2 & 22.70 & 102.1 & 0.551 \\
CUIG-GradProj & {\underline{15.01}} & {\underline{384.6}} & {\underline{0.851}} & \bfseries 0.0 & 14.77 & 380.9 & 0.856 & 0.0 & 13.86 & 313.2 & 0.848 \\
Sculpting Memory & 17.14 & 165.4 & 0.645 & {\underline{0.6}} & 17.66 & 153.9 & 0.640 & 0.6 & 25.51 & {\underline{45.3}} & 0.418 \\
\midrule
Ours & 21.59 & 76.9 & 0.456 & 4.0 & \bfseries 31.80 & \bfseries 35.4 & \bfseries 0.237 & \bfseries 65.4 & \bfseries 26.31 & 46.2 & {\underline{0.386}} \\
\bottomrule
\end{tabular}
}
\end{table}

\begin{table}[h]
\centering
\caption{Style erasure after continually erasing 100 artistic styles on SD v1.5. Bold and \underline{underline} mark the best and second-best edited results.}
\label{tab:sd15_style}
\resizebox{\hsize}{!}{
\renewcommand{\arraystretch}{1.20}
\begin{tabular}{l|CFLE|CFLE|CFL}
\toprule
 & \multicolumn{4}{c|}{Erase set} & \multicolumn{4}{c|}{Retain set} & \multicolumn{3}{c}{MS-COCO} \\
 & {CS $\downarrow$} & {FID $\uparrow$} & {LPIPS $\uparrow$} & {Aes $\uparrow$} & {CS $\uparrow$} & {FID $\downarrow$} & {LPIPS $\downarrow$} & {Aes $\uparrow$} & {CS $\uparrow$} & {FID $\downarrow$} & {LPIPS $\downarrow$} \\
\midrule
SD v1.5 & 28.34 & 0.0 & 0.000 & 6.01 & 28.24 & 0.0 & 0.000 & 6.00 & 26.43 & 0.0 & 0.000 \\
\midrule
ESD & 24.50 & 50.9 & 0.485 & {\underline{5.76}} & {\underline{24.53}} & {\underline{49.4}} & 0.485 & {\underline{5.77}} & {\underline{25.71}} & {\underline{42.9}} & 0.388 \\
UCE & 17.09 & 227.8 & 0.860 & 4.59 & 15.41 & 182.4 & 0.795 & 4.24 & 13.70 & 210.2 & 0.776 \\
ConAbl & 23.70 & 66.4 & 0.468 & 5.67 & 24.01 & 61.3 & {\underline{0.468}} & 5.66 & 25.37 & \bfseries 40.6 & \bfseries 0.355 \\
RECE & \bfseries 14.60 & \bfseries 663.7 & \bfseries 0.996 & 4.60 & 14.05 & 668.9 & 0.997 & 4.60 & 16.28 & 610.4 & 1.002 \\
SPEED & {\underline{15.81}} & 225.3 & 0.724 & 4.83 & 17.69 & 170.1 & 0.688 & 5.02 & 14.31 & 186.7 & 0.655 \\
CUIG-GradProj & 16.55 & {\underline{284.7}} & {\underline{0.885}} & 4.00 & 16.02 & 280.0 & 0.884 & 4.03 & 19.95 & 143.7 & 0.740 \\
Sculpting Memory & 21.45 & 111.1 & 0.597 & 5.44 & 21.64 & 104.2 & 0.590 & 5.43 & 25.20 & 43.0 & 0.402 \\
\midrule
Ours & 23.90 & 90.6 & 0.514 & \bfseries 5.87 & \bfseries 27.17 & \bfseries 37.9 & \bfseries 0.380 & \bfseries 5.92 & \bfseries 26.23 & 44.0 & {\underline{0.370}} \\
\bottomrule
\end{tabular}
}
\end{table}

\begin{table}[h]
\centering
\caption{Instance erasure after continually erasing 100 instances on SD v1.5. Bold and \underline{underline} mark the best and second-best edited results.}
\label{tab:sd15_instance}
\resizebox{\hsize}{!}{
\renewcommand{\arraystretch}{1.20}
\begin{tabular}{l|CFLA|CFLA|CFL}
\toprule
 & \multicolumn{4}{c|}{Erase set} & \multicolumn{4}{c|}{Retain set} & \multicolumn{3}{c}{MS-COCO} \\
 & {CS $\downarrow$} & {FID $\uparrow$} & {LPIPS $\uparrow$} & {Acc $\downarrow$} & {CS $\uparrow$} & {FID $\downarrow$} & {LPIPS $\downarrow$} & {Acc $\uparrow$} & {CS $\uparrow$} & {FID $\downarrow$} & {LPIPS $\downarrow$} \\
\midrule
SD v1.5 & 28.05 & 0.0 & 0.000 & 79.9 & 28.52 & 0.0 & 0.000 & 78.9 & 26.43 & 0.0 & 0.000 \\
\midrule
ESD & 24.28 & 38.9 & 0.535 & 58.6 & 24.60 & 36.0 & {\underline{0.517}} & {\underline{58.0}} & 24.50 & \bfseries 46.6 & 0.440 \\
UCE & \bfseries 14.97 & {\underline{301.0}} & 0.809 & 1.6 & 15.25 & 247.8 & 0.791 & 0.2 & 14.03 & 264.1 & 0.758 \\
ConAbl & 19.76 & 74.4 & 0.567 & 15.8 & 20.36 & 71.2 & 0.562 & 18.8 & {\underline{24.89}} & {\underline{46.8}} & \bfseries 0.411 \\
RECE & 17.23 & \bfseries 606.3 & \bfseries 1.006 & \bfseries 0.0 & 17.12 & 594.9 & 1.006 & 1.0 & 16.28 & 610.4 & 1.002 \\
SPEED & 17.59 & 89.5 & 0.673 & 5.0 & {\underline{25.23}} & {\underline{26.9}} & 0.534 & 57.6 & 21.73 & 86.7 & 0.594 \\
CUIG-GradProj & 16.78 & 256.4 & {\underline{0.913}} & {\underline{1.1}} & 16.64 & 250.0 & 0.910 & 0.4 & 13.48 & 234.8 & 0.914 \\
Sculpting Memory & {\underline{15.48}} & 149.7 & 0.703 & 3.1 & 15.48 & 155.2 & 0.698 & 3.0 & 22.48 & 63.9 & 0.526 \\
\midrule
Ours & 21.37 & 46.2 & 0.570 & 16.0 & \bfseries 27.80 & \bfseries 17.6 & \bfseries 0.372 & \bfseries 75.6 & \bfseries 25.69 & 53.5 & {\underline{0.417}} \\
\bottomrule
\end{tabular}
}
\end{table}

\subsection{Backbone Generalization to Stable Diffusion v2.1}
\label{app:sd21}

Stable Diffusion v2.1 replaces the CLIP-L text encoder with the 1024-dimensional OpenCLIP-H and is the hardest tested backbone: every method erases more weakly, and the margins narrow accordingly (Tables~\ref{tab:sd21_celebrity}, \ref{tab:sd21_style}, and \ref{tab:sd21_instance}). CEASE still improves the retain and COCO preservation metrics over SPEED on all three tasks, whereas some baselines preserve well only because they barely erase: UCE leaves most of the erased celebrities recognizable, and ESD leaves the erased instances nearly intact. Overall, the transfer study supports a consistent preservation advantage over the underlying SPEED solver across both text encoders, at the cost of shallower erasure on several tasks.

\begin{table}[h]
\centering
\caption{Celebrity erasure after continually erasing 100 celebrities on SD v2.1. Bold and \underline{underline} mark the best and second-best edited results.}
\label{tab:sd21_celebrity}
\resizebox{\hsize}{!}{
\renewcommand{\arraystretch}{1.20}
\begin{tabular}{l|CFLA|CFLA|CFL}
\toprule
 & \multicolumn{4}{c|}{Erase set} & \multicolumn{4}{c|}{Retain set} & \multicolumn{3}{c}{MS-COCO} \\
 & {CS $\downarrow$} & {FID $\uparrow$} & {LPIPS $\uparrow$} & {Acc $\downarrow$} & {CS $\uparrow$} & {FID $\downarrow$} & {LPIPS $\downarrow$} & {Acc $\uparrow$} & {CS $\uparrow$} & {FID $\downarrow$} & {LPIPS $\downarrow$} \\
\midrule
SD v2.1 & 30.60 & 0.0 & 0.000 & 88.8 & 31.03 & 0.0 & 0.000 & 75.6 & 26.32 & 0.0 & 0.000 \\
\midrule
ESD & 25.85 & 103.5 & 0.520 & 62.0 & 25.48 & 109.2 & 0.523 & 41.8 & 23.43 & 52.8 & 0.481 \\
UCE & 27.82 & 47.1 & 0.398 & 47.0 & 30.74 & 42.2 & 0.362 & 68.2 & 26.11 & 48.5 & 0.442 \\
ConAbl & 15.63 & 214.3 & 0.603 & \bfseries 0.0 & 15.68 & 206.1 & 0.604 & 0.0 & 24.60 & 47.9 & 0.451 \\
RECE & 16.51 & \bfseries 549.3 & \bfseries 0.973 & \bfseries 0.0 & 16.27 & 562.3 & 0.971 & 0.0 & 16.28 & 614.3 & 1.002 \\
SPEED & 24.90 & 53.7 & 0.377 & {\underline{12.2}} & {\underline{31.00}} & {\underline{32.8}} & {\underline{0.245}} & \bfseries 72.0 & {\underline{26.12}} & {\underline{41.9}} & {\underline{0.340}} \\
CUIG-GradProj & \bfseries 14.98 & {\underline{381.1}} & {\underline{0.864}} & \bfseries 0.0 & 14.71 & 381.4 & 0.864 & 0.0 & 12.50 & 284.0 & 0.954 \\
Sculpting Memory & {\underline{15.15}} & 164.2 & 0.693 & \bfseries 0.0 & 15.26 & 152.3 & 0.680 & 0.4 & 24.48 & 58.6 & 0.527 \\
\midrule
Ours & 26.91 & 38.8 & 0.326 & 34.2 & \bfseries 31.01 & \bfseries 31.2 & \bfseries 0.217 & {\underline{69.6}} & \bfseries 26.14 & \bfseries 38.2 & \bfseries 0.290 \\
\bottomrule
\end{tabular}
}
\end{table}

\begin{table}[h]
\centering
\caption{Style erasure after continually erasing 100 artistic styles on SD v2.1. Bold and \underline{underline} mark the best and second-best edited results.}
\label{tab:sd21_style}
\resizebox{\hsize}{!}{
\renewcommand{\arraystretch}{1.20}
\begin{tabular}{l|CFLE|CFLE|CFL}
\toprule
 & \multicolumn{4}{c|}{Erase set} & \multicolumn{4}{c|}{Retain set} & \multicolumn{3}{c}{MS-COCO} \\
 & {CS $\downarrow$} & {FID $\uparrow$} & {LPIPS $\uparrow$} & {Aes $\uparrow$} & {CS $\uparrow$} & {FID $\downarrow$} & {LPIPS $\downarrow$} & {Aes $\uparrow$} & {CS $\uparrow$} & {FID $\downarrow$} & {LPIPS $\downarrow$} \\
\midrule
SD v2.1 & 27.26 & 0.0 & 0.000 & 6.01 & 27.37 & 0.0 & 0.000 & 5.99 & 26.32 & 0.0 & 0.000 \\
\midrule
ESD & 25.71 & 41.6 & 0.400 & 5.95 & 25.69 & 41.8 & 0.399 & 5.94 & 25.45 & 41.4 & 0.366 \\
UCE & 25.89 & 28.7 & 0.316 & {\underline{5.99}} & {\underline{26.08}} & {\underline{28.0}} & {\underline{0.303}} & {\underline{5.97}} & {\underline{25.91}} & {\underline{40.3}} & {\underline{0.339}} \\
ConAbl & 25.25 & 43.2 & 0.424 & 5.84 & 25.49 & 40.9 & 0.415 & 5.83 & 25.51 & 41.0 & 0.350 \\
RECE & \bfseries 12.45 & {\underline{442.0}} & {\underline{0.866}} & 3.48 & 12.14 & 435.4 & 0.862 & 3.43 & 14.47 & 403.6 & 0.878 \\
SPEED & 24.93 & 40.7 & 0.404 & 5.84 & 25.29 & 37.9 & 0.370 & 5.89 & 25.35 & 50.3 & 0.396 \\
CUIG-GradProj & {\underline{14.10}} & \bfseries 468.9 & \bfseries 0.889 & 3.39 & 13.55 & 462.9 & 0.886 & 3.40 & 13.74 & 392.9 & 0.794 \\
Sculpting Memory & 19.42 & 98.3 & 0.609 & 5.46 & 20.00 & 91.6 & 0.599 & 5.47 & 24.57 & 47.3 & 0.453 \\
\midrule
Ours & 26.14 & 25.6 & 0.266 & \bfseries 6.01 & \bfseries 26.41 & \bfseries 24.7 & \bfseries 0.247 & \bfseries 6.01 & \bfseries 25.97 & \bfseries 37.7 & \bfseries 0.291 \\
\bottomrule
\end{tabular}
}
\end{table}

\begin{table}[h]
\centering
\caption{Instance erasure after continually erasing 100 instances on SD v2.1. Bold and \underline{underline} mark the best and second-best edited results.}
\label{tab:sd21_instance}
\resizebox{\hsize}{!}{
\renewcommand{\arraystretch}{1.20}
\begin{tabular}{l|CFLA|CFLA|CFL}
\toprule
 & \multicolumn{4}{c|}{Erase set} & \multicolumn{4}{c|}{Retain set} & \multicolumn{3}{c}{MS-COCO} \\
 & {CS $\downarrow$} & {FID $\uparrow$} & {LPIPS $\uparrow$} & {Acc $\downarrow$} & {CS $\uparrow$} & {FID $\downarrow$} & {LPIPS $\downarrow$} & {Acc $\uparrow$} & {CS $\uparrow$} & {FID $\downarrow$} & {LPIPS $\downarrow$} \\
\midrule
SD v2.1 & 26.39 & 0.0 & 0.000 & 74.3 & 26.43 & 0.0 & 0.000 & 70.2 & 26.32 & 0.0 & 0.000 \\
\midrule
ESD & 25.64 & 21.6 & 0.395 & 69.2 & 25.51 & 19.8 & {\underline{0.386}} & 64.4 & {\underline{25.72}} & \bfseries 40.1 & {\underline{0.347}} \\
UCE & 17.50 & 125.5 & 0.666 & 5.8 & 19.15 & 76.9 & 0.617 & 19.8 & 18.43 & 135.4 & 0.612 \\
ConAbl & 20.67 & 60.4 & 0.567 & 27.2 & 20.72 & 60.5 & 0.567 & 27.0 & 25.17 & 45.4 & 0.396 \\
RECE & {\underline{16.66}} & \bfseries 351.8 & \bfseries 0.883 & {\underline{1.0}} & 16.44 & 327.3 & 0.874 & 0.0 & 13.35 & 419.1 & 0.880 \\
SPEED & 22.30 & 36.4 & 0.507 & 34.6 & {\underline{25.73}} & {\underline{19.0}} & 0.403 & {\underline{65.7}} & 25.55 & 51.2 & 0.408 \\
CUIG-GradProj & 17.37 & {\underline{350.0}} & {\underline{0.822}} & \bfseries 0.9 & 16.79 & 351.9 & 0.828 & 0.0 & 13.40 & 351.6 & 0.839 \\
Sculpting Memory & \bfseries 14.43 & 178.6 & 0.703 & 2.5 & 14.39 & 179.5 & 0.702 & 4.7 & 18.30 & 119.7 & 0.615 \\
\midrule
Ours & 24.04 & 25.1 & 0.445 & 47.7 & \bfseries 26.01 & \bfseries 16.2 & \bfseries 0.338 & \bfseries 67.1 & \bfseries 25.85 & {\underline{43.1}} & \bfseries 0.340 \\
\bottomrule
\end{tabular}
}
\end{table}

\subsection{Order Robustness}
\label{app:order}

We rerun the 100-step celebrity sequence under five random permutations of the concept order with editing seed 0. Table~\ref{tab:order} reports the per-order results: the variation across orders is small on every metric relative to the between-method differences in the main tables.

\begin{table}[h]
\centering
\caption{Erasure-order robustness on the celebrity task: mean\,$\pm$\,std over five random orderings of the 100 targets, using five templates and one image-sampling seed per template. The editing seed is fixed at 0, and CEASE includes the same retain augmentation as in the main table.}
\label{tab:order}
\resizebox{\hsize}{!}{
\renewcommand{\arraystretch}{1.25}
\begin{tabular}{l|ccc|ccc|ccc}
\toprule
 & \multicolumn{3}{c|}{Erase set} & \multicolumn{3}{c|}{Retain set} & \multicolumn{3}{c}{MS-COCO} \\
 & {CS $\downarrow$} & {FID $\uparrow$} & {LPIPS $\uparrow$} & {CS $\uparrow$} & {FID $\downarrow$} & {LPIPS $\downarrow$} & {CS $\uparrow$} & {FID $\downarrow$} & {LPIPS $\downarrow$} \\
\midrule
Ours & 21.35{\scriptsize\,$\pm$0.06} & 81.5{\scriptsize\,$\pm$0.8} & 0.543{\scriptsize\,$\pm$0.002} & 31.57{\scriptsize\,$\pm$0.08} & 48.9{\scriptsize\,$\pm$1.0} & 0.428{\scriptsize\,$\pm$0.003} & 26.22{\scriptsize\,$\pm$0.05} & 46.3{\scriptsize\,$\pm$0.7} & 0.387{\scriptsize\,$\pm$0.009} \\
\bottomrule
\end{tabular}
}
\end{table}

\subsection{Runtime Efficiency}
\label{app:runtime}

\begin{figure}[h]
\centering
\includegraphics[width=0.9\linewidth]{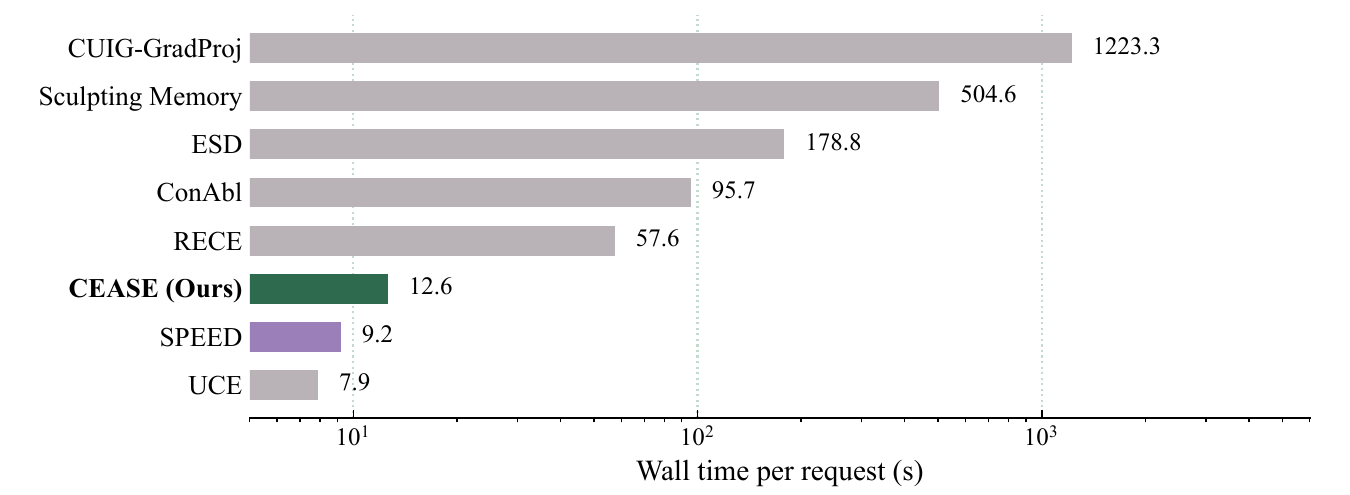}
\caption{Single-request wall time on one H100 GPU. CEASE stays in the seconds-level regime of the closed-form methods, one to two orders of magnitude below the training-based baselines.}
\label{fig:runtime}
\end{figure}

Fig.~\ref{fig:runtime} compares single-request wall time on one H100 GPU. CEASE completes a request in about 13 seconds, staying in the seconds-level regime of the closed-form methods, whereas the training-based methods take minutes per request, and this gap compounds over the continual sequence. Relative to SPEED, AIC and HOC add about $2\%$ to the edit loop.

\subsection{Additional Qualitative Comparisons}
\label{app:qual_v14}

\begin{figure}[tp]
\centering
\includegraphics[width=\linewidth,height=0.86\textheight,keepaspectratio]{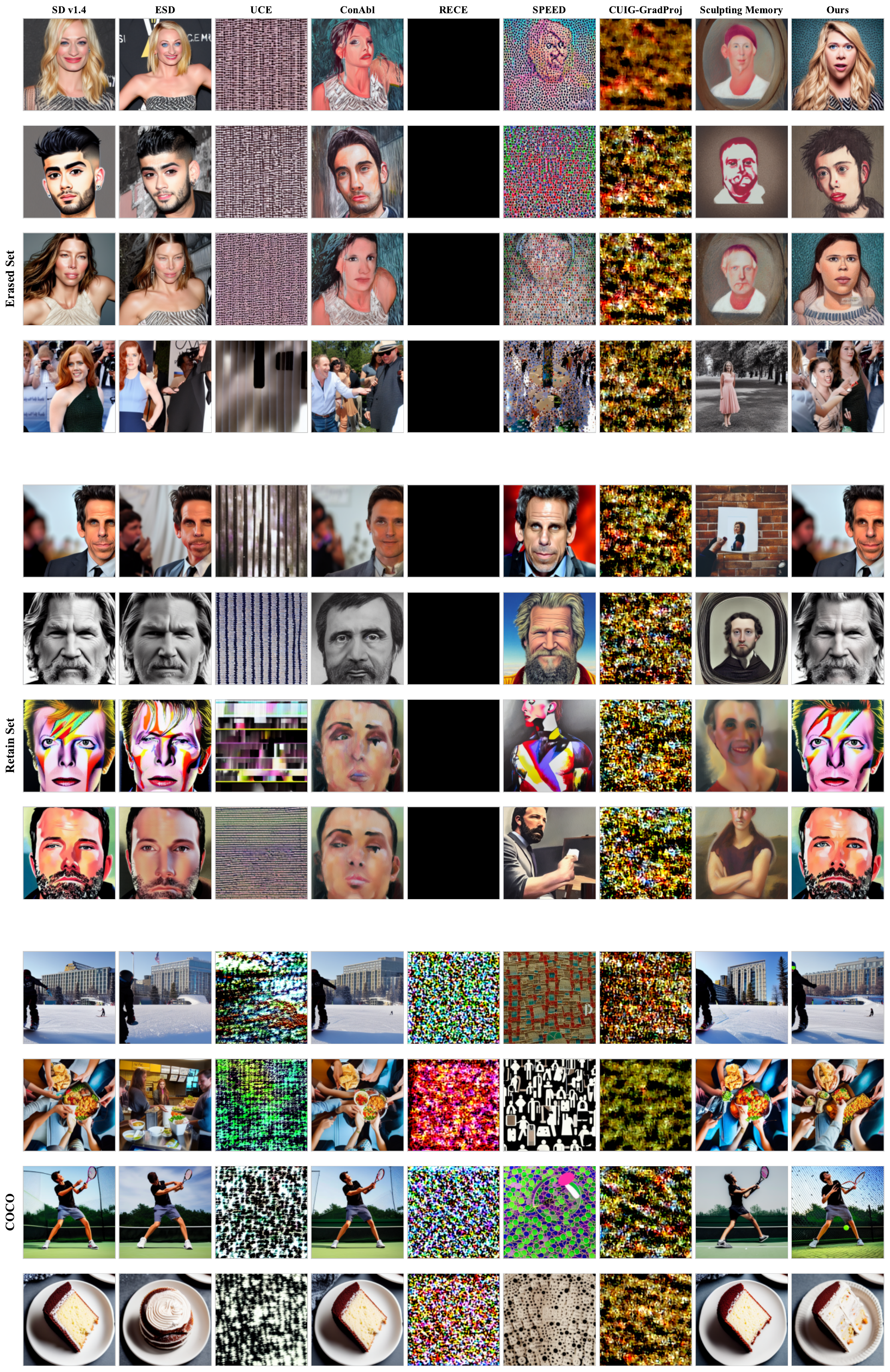}
\caption{Additional celebrity examples after the full 100-step sequence on Stable Diffusion v1.4.}
\label{fig:qual_v14_extra}
\end{figure}

\begin{figure}[tp]
\centering
\includegraphics[width=\linewidth,height=0.86\textheight,keepaspectratio]{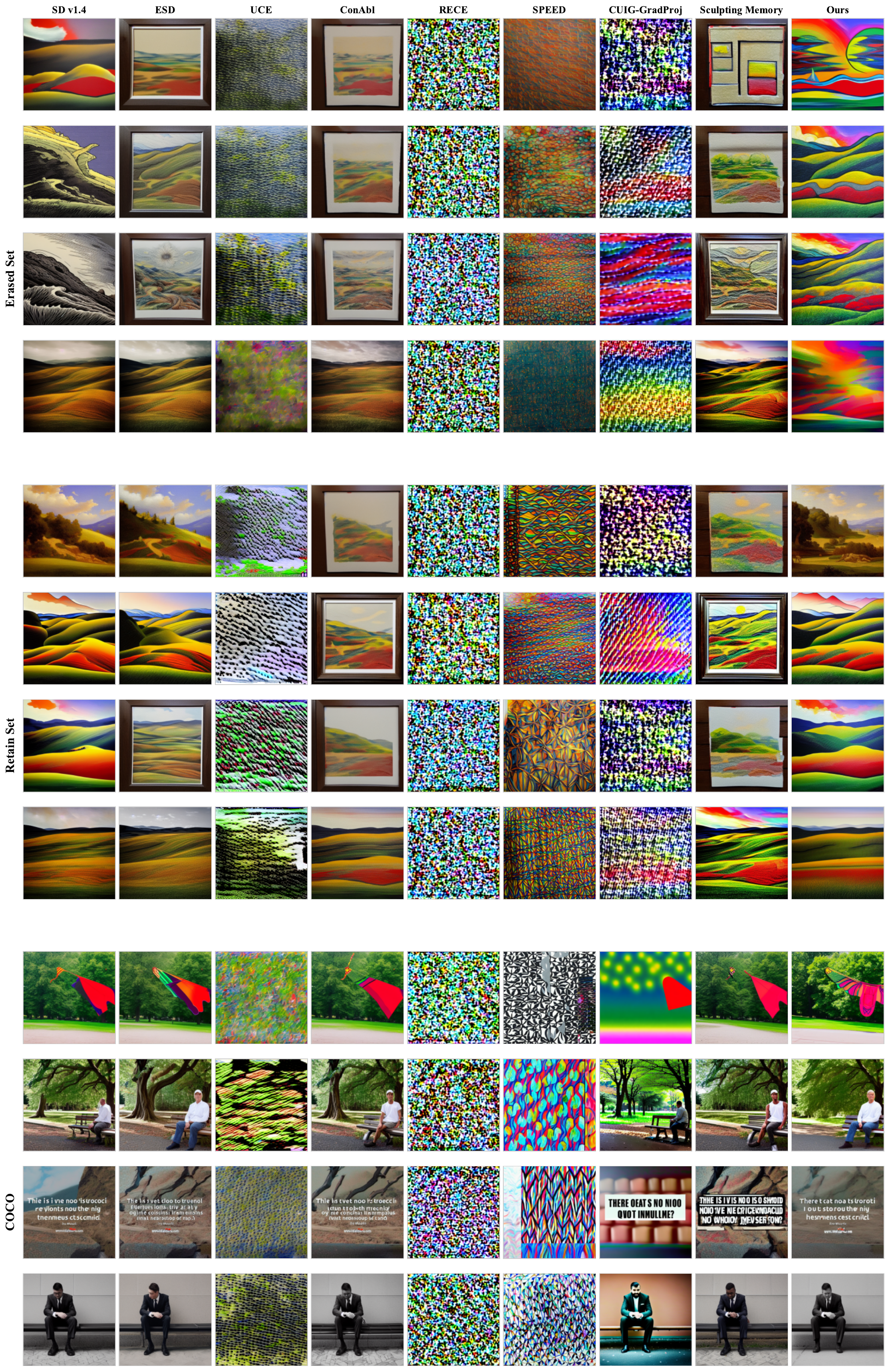}
\caption{Additional style examples after the full 100-step sequence on Stable Diffusion v1.4.}
\label{fig:qual_style}
\end{figure}

\begin{figure}[tp]
\centering
\includegraphics[width=\linewidth,height=0.86\textheight,keepaspectratio]{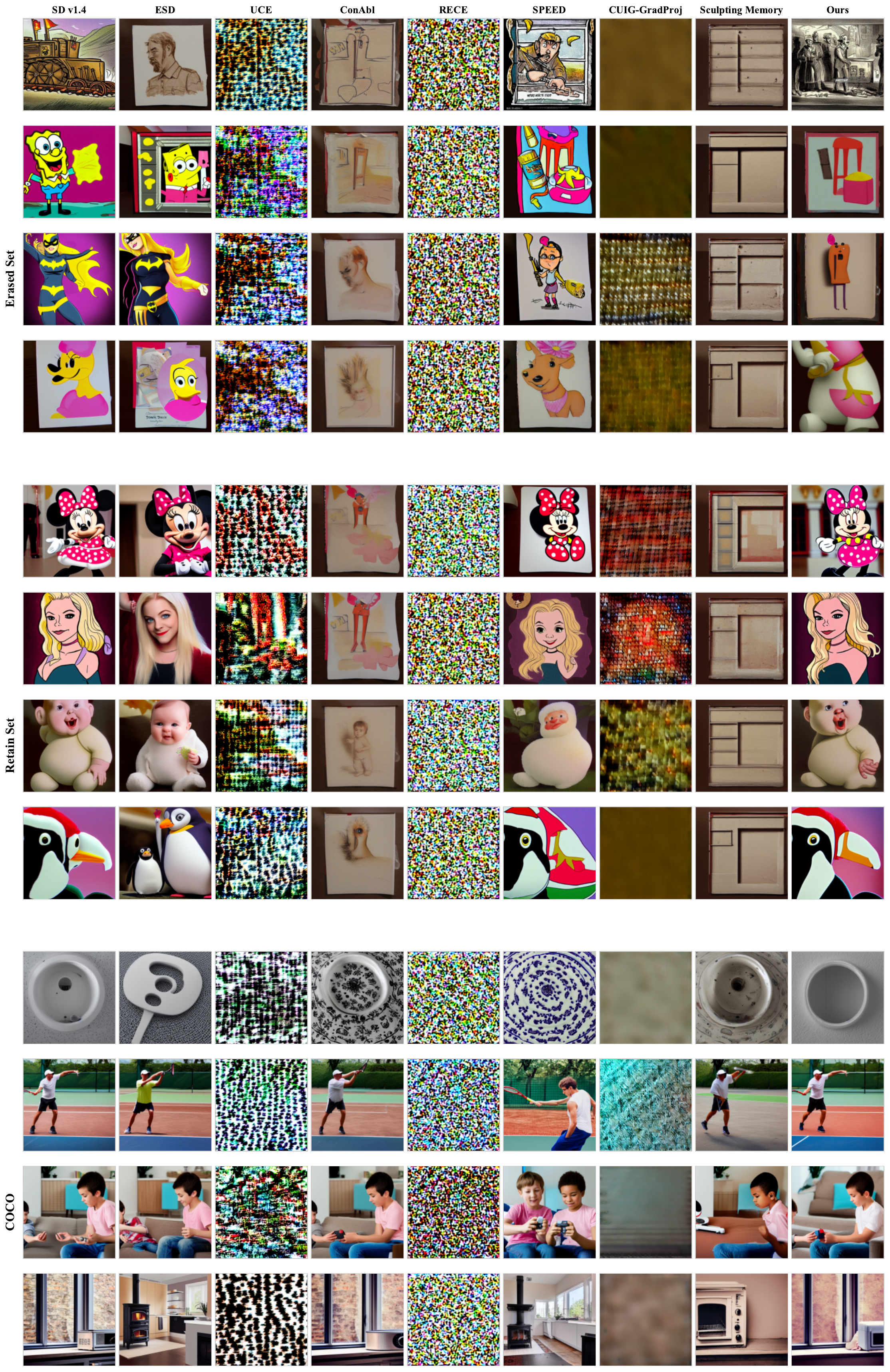}
\caption{Additional instance examples after the full 100-step sequence on Stable Diffusion v1.4.}
\label{fig:qual_instance}
\end{figure}

\begin{figure}[tp]
\centering
\includegraphics[width=\linewidth,height=0.86\textheight,keepaspectratio]{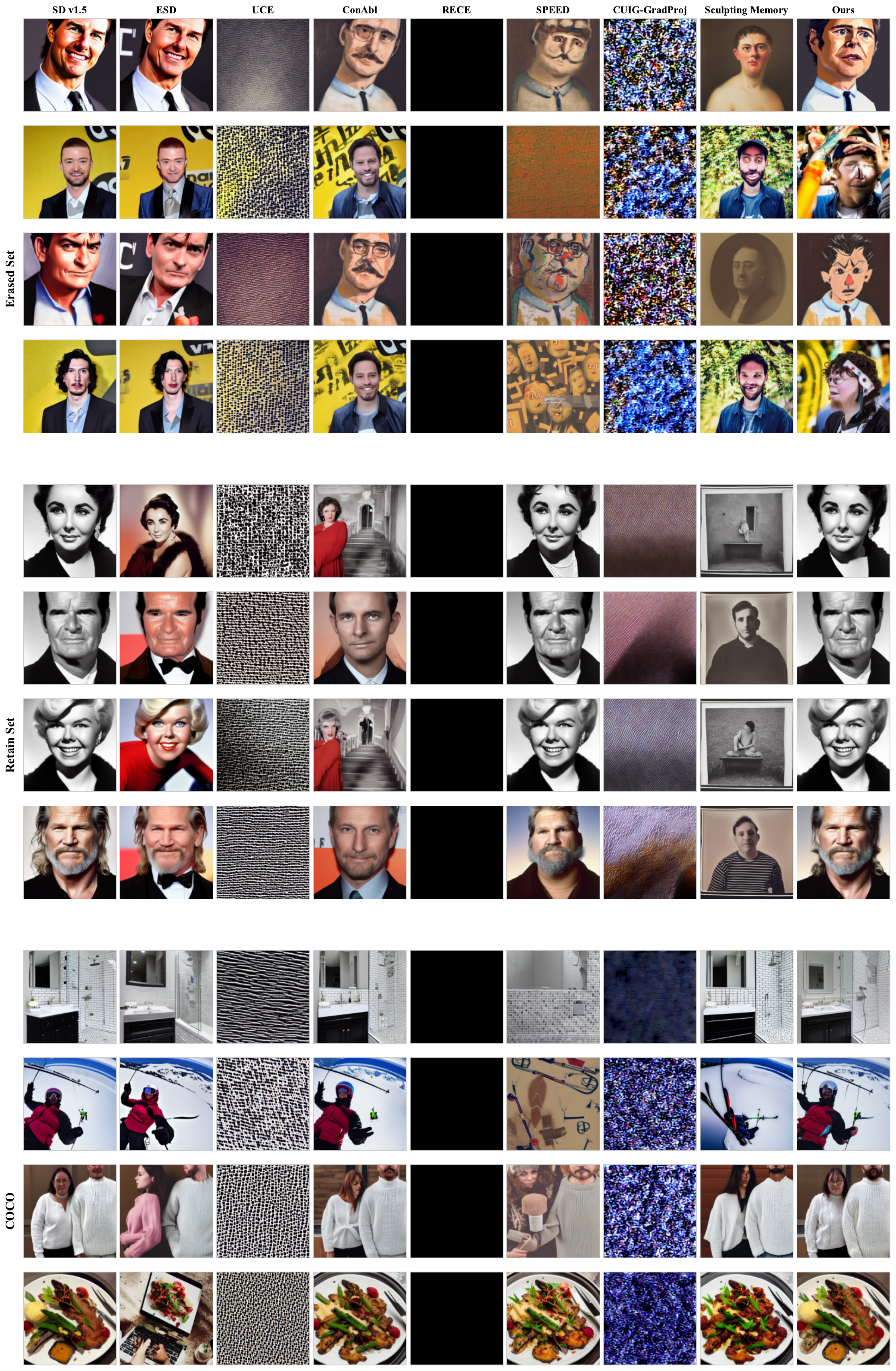}
\caption{Celebrity examples after the full 100-step sequence on Stable Diffusion v1.5.}
\label{fig:qual_sd15}
\end{figure}

\begin{figure}[tp]
\centering
\includegraphics[width=\linewidth,height=0.86\textheight,keepaspectratio]{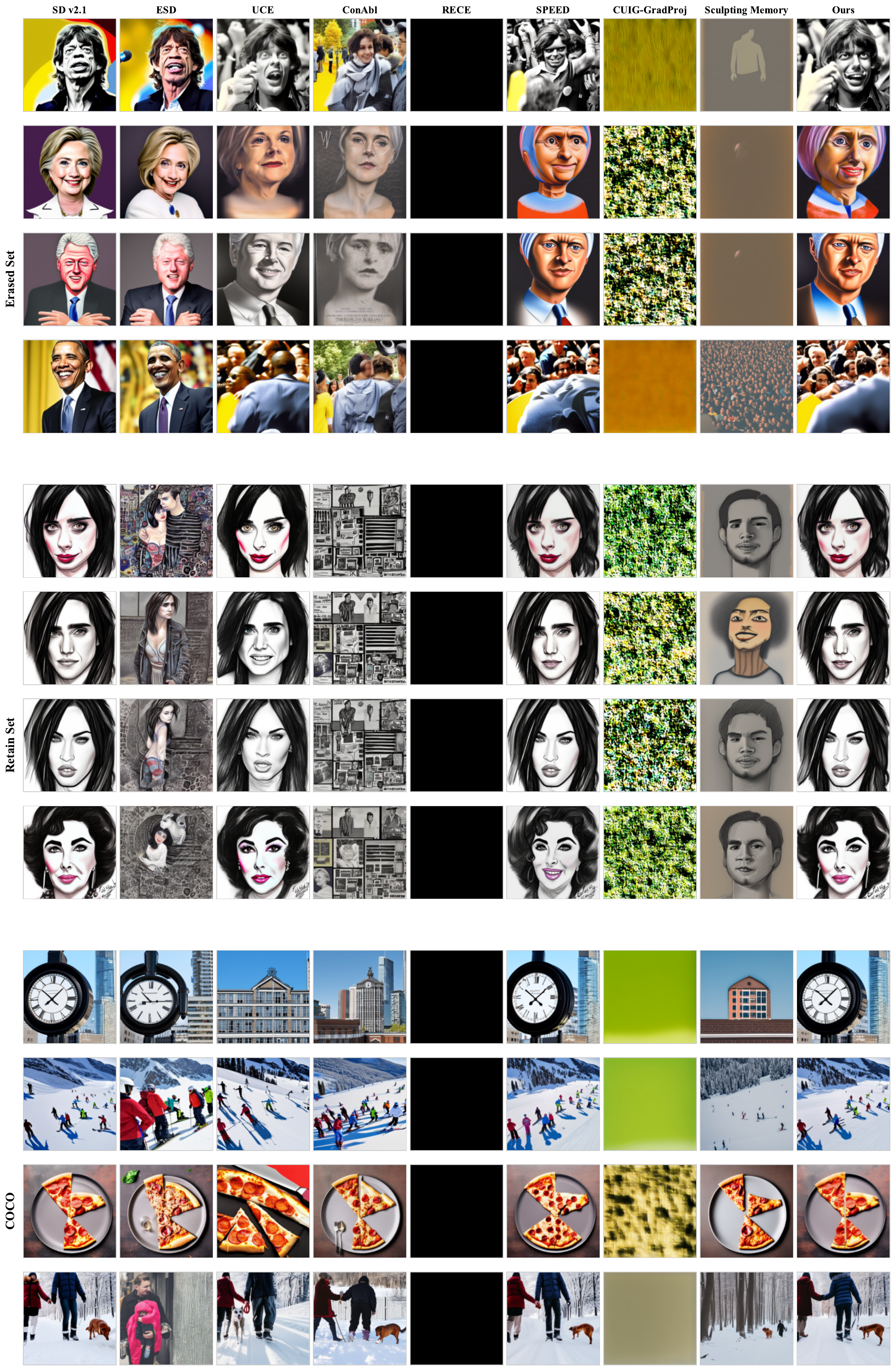}
\caption{Celebrity examples after the full 100-step sequence on Stable Diffusion v2.1.}
\label{fig:qual_sd21}
\end{figure}

This section provides qualitative comparisons beyond the single case shown in the main text. Fig.~\ref{fig:qual_v14_extra} extends the main-text comparison to four cases per category on the celebrity task, Figs.~\ref{fig:qual_style} and \ref{fig:qual_instance} give the same comparison on the style and instance tasks, and Figs.~\ref{fig:qual_sd15} and \ref{fig:qual_sd21} repeat the celebrity comparison on the Stable Diffusion v1.5 and v2.1 backbones. Every figure follows the same layout: columns show the unedited reference followed by ESD, UCE, ConAbl, RECE, SPEED, CUIG-GradProj, Sculpting Memory, and CEASE after the full 100-step sequence, and the three row blocks show four cases each of erased targets, retained concepts, and unrelated MS-COCO prompts.

The failure modes match the quantitative results and the main-text analysis. On the erased rows, CEASE redirects each target to a coherent anchor substitute, whereas ESD often leaves the erased identity or style still recognizable, and RECE, UCE, and CUIG-GradProj collapse to noise-like textures once the sequence is long. Sculpting Memory remains photorealistic but drifts to subjects unrelated to either the target or its anchor. On the retained and COCO rows, the picture inverts: the destructive baselines corrupt concepts they were never asked to erase, SPEED preserves some cases but degrades others as its accumulated interference grows, and CEASE stays visually closest to the reference across both row blocks. The v1.5 and v2.1 figures show the same pattern, confirming that these behaviors are properties of the continual setting rather than of the v1.4 backbone.

\clearpage

\clearpage

\clearpage

\clearpage

\clearpage

\clearpage
\subsection{Scope and Protocol of the Expanded Noise Comparisons}
\label{app:noise_scope}

\begin{figure}[ht]
\centering
\includegraphics[width=\linewidth]{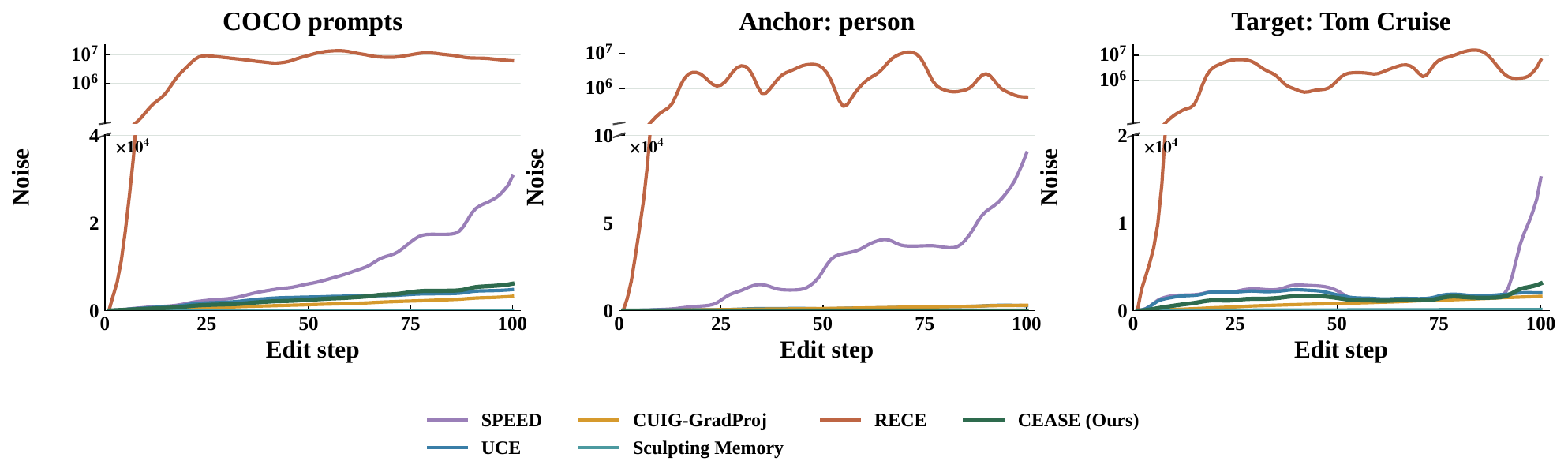}
\caption{Completed candidate comparisons on the common value-projection diagnostic, including methods with lower noise than CEASE. The one-epoch RECE configuration and UCE sequence are defined above.}
\label{fig:noise_candidates}
\end{figure}

The expanded diagnostic comparison uses the same 100-celebrity order, frozen SD v1.4 text encoder, and 16 cross-attention value projections for every method. It does not measure changes to key projections or other trainable parameters. Every baseline in Figs.~\ref{fig:main} and~\ref{fig:ablation_scaling} also appears in Table~\ref{tab:celebrity}. Fig.~\ref{fig:noise_candidates} adds CUIG-GradProj and Sculpting Memory to those comparisons. Several methods have smaller value-projection noise than CEASE despite weaker generation quality; this diagnostic therefore cannot be used alone to rank arbitrary erasure methods.

\begin{figure}[ht]
\centering
\includegraphics[width=\linewidth]{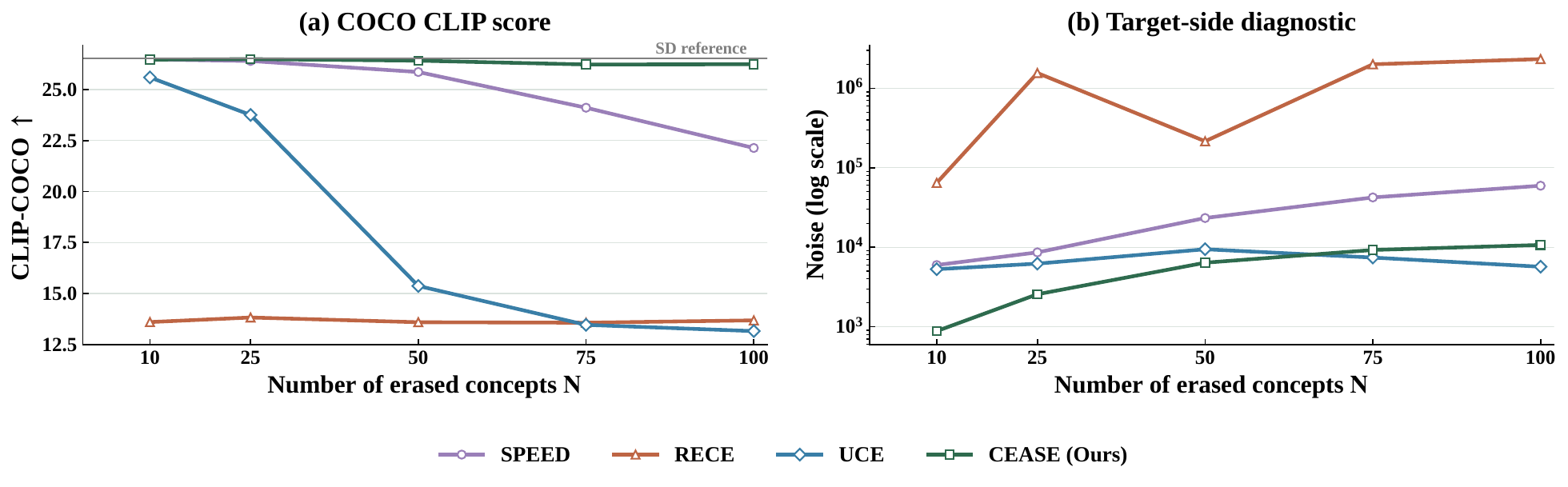}
\caption{Scaling on the celebrity benchmark at $N\in\{10,25,50,75,100\}$. (a) COCO CLIP score on 1,000 prompts; the gray line is the unedited SD reference. (b) Target-averaged value-projection diagnostic on a logarithmic axis. Markers denote measured checkpoints; COCO FID at the same checkpoints is shown in Fig.~\ref{fig:main}(a).}
\label{fig:scaling_supplement}
\end{figure}

\noindent\textbf{Definitions.} Fig.~\ref{fig:ablation_scaling} uses the fixed-prompt diagnostic
\begin{equation}
 D_N(c)=\sum_\ell\|(W_{N,\ell}-W_{0,\ell})c\|_2^2
       -\sum_\ell\|(W_{N,\ell}-W_{N-1,\ell})c\|_2^2.
\end{equation}
The COCO curve averages $D_N$ over 1,000 caption embeddings; the other curves use the last non-EOS token of ``person'' and ``Tom Cruise in an official photo''. This operational diagnostic subtracts the latest request's contribution. Fig.~\ref{fig:scaling_supplement}(b), in contrast, averages over the erased concepts seen so far and subtracts each concept's own request, $\frac1N\sum_{i\leq N}[\sum_\ell\|(W_{N,\ell}-W_{0,\ell})c_i\|_2^2-\sum_\ell\|(W_{i,\ell}-W_{i-1,\ell})c_i\|_2^2]$. SPEED and CEASE use signed values without absolute-value conversion; all other methods use the absolute value of each recorded diagnostic, taken after averaging where applicable. The measurement files retain the signed differences of energies. Original measurement files remain unchanged. These diagnostics alone do not determine erasure or preservation quality.

\noindent\textbf{RECE diagnostic run.} The added curves use one adversarial epoch per request, with erase scale 1.0, preserve scale 0.1, $\lambda=0.1$, adversarial regularization 0.1, and the replace objective. This is labeled ``RECE'' in the figures and is separate from the three-epoch configuration described above. A three-epoch rerun failed during request 61 with a singular-matrix error and is not presented as a complete trajectory. We cache only the frozen retain embeddings and omit diagnostic preview images; a one-request check found all 32 saved key/value tensors bitwise identical to the uncached implementation. Every plotted milestone and qualitative image comes from the complete one-epoch rerun, without joining it to a historical final checkpoint.

\noindent\textbf{UCE sequence and shared sampling protocol.} UCE uses its saved 100-step celebrity sequence with erase and preserve scales of 1.0, $\lambda=0.5$, and the shared replacement ``person''. All 32 final key/value tensors match the checkpoint used for the UCE row in Table~\ref{tab:celebrity}. Its noise curve uses all 100 checkpoints; COCO metrics are measured at $N\in\{10,25,50,75,100\}$. Every plotted method uses FP16 sampling with 20 DPM-Solver steps, guidance 7.5, COCO seed 0, batch size 10, and the same 1,000 prompts. FID uses the same unedited SD image set as the existing curves. Figure~\ref{fig:main}(a) shows point estimates; no uncertainty band is drawn without measured bounds.

\begin{figure}[ht]
\centering
\includegraphics[width=\linewidth,height=0.84\textheight,keepaspectratio]{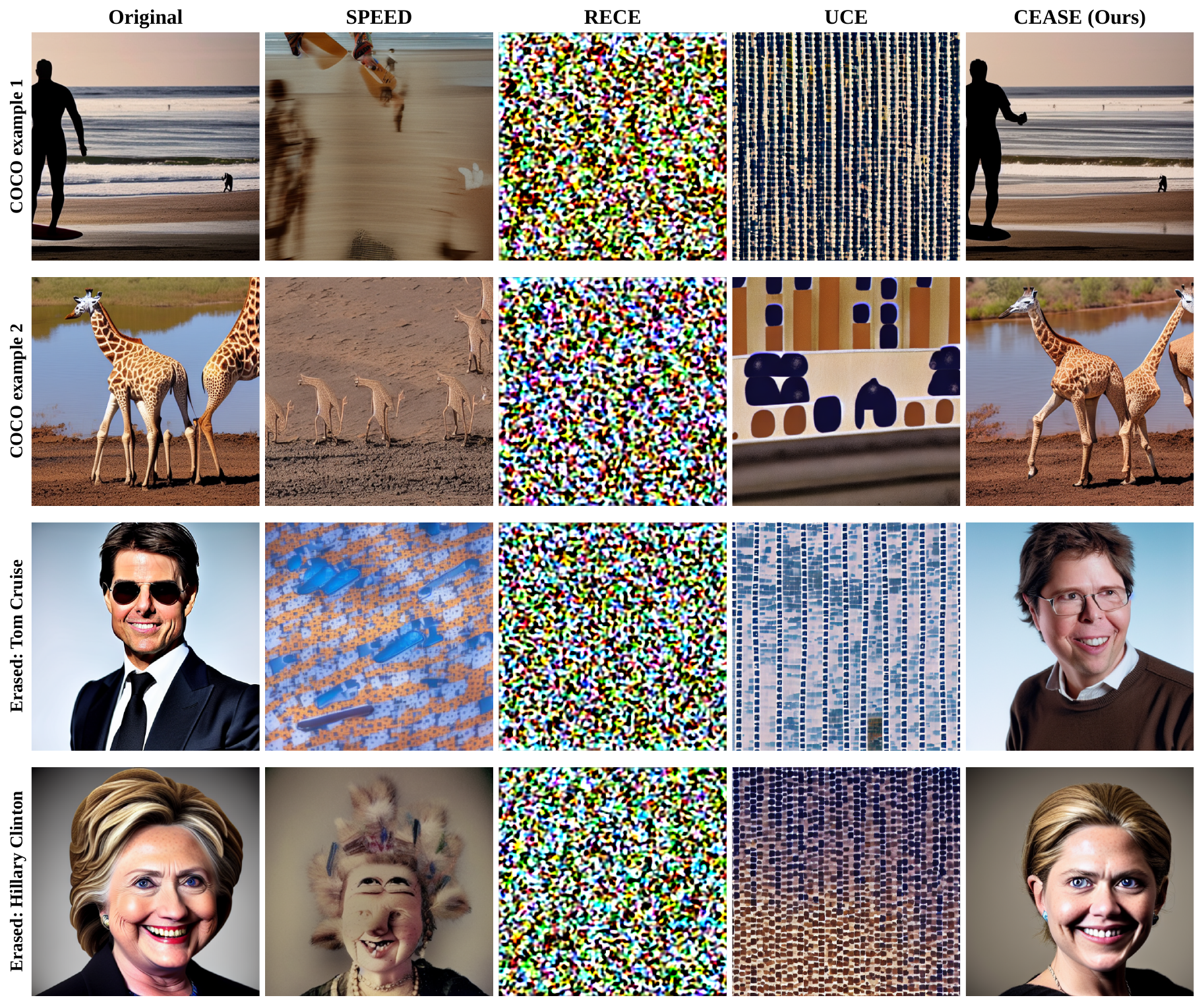}
\caption{COCO and erased-concept comparisons for SPEED, RECE, UCE, and CEASE after 100 celebrity erasures. Every row uses the same prompt and random seed with FP16 sampling, and erased-target sampling retains the original batch positions. RECE uses one adversarial epoch per request.}
\label{fig:qual_noise_methods}
\end{figure}

\end{document}